%% file: main.tex
\documentclass{article}

\usepackage[preprint]{neurips_2026}

\usepackage[utf8]{inputenc} % allow utf-8 input
\usepackage[T1]{fontenc}    % use 8-bit T1 fonts
\usepackage{hyperref}       % hyperlinks
\usepackage{url}            % simple URL typesetting
\usepackage{booktabs}       % professional-quality tables
\usepackage{graphicx}
\usepackage{amsfonts}       % blackboard math symbols
\usepackage{amssymb}
\usepackage{nicefrac}       % compact symbols for 1/2, etc.
\usepackage{microtype}      % microtypography
\usepackage{xcolor}         % colors
\usepackage[capitalize]{cleveref}
\usepackage{algorithm}
\usepackage{algpseudocode}
\usepackage{listings}
\usepackage[normalem]{ulem}
\usepackage{tikz}
\usetikzlibrary{arrows.meta,positioning,calc,shapes.geometric,decorations.pathmorphing,decorations.pathreplacing}

\algrenewcommand\algorithmicrequire{\textbf{Input:}}
\algrenewcommand\algorithmicensure{\textbf{Output:}}
\algtext*{EndFor}
\algtext*{EndIf}

\definecolor{qualboxborder}{RGB}{214,214,214}
\definecolor{qualaccent}{RGB}{72,96,129}
\definecolor{qualtagbg}{RGB}{244,246,249}
\definecolor{qualsembg}{RGB}{244,246,249}
\definecolor{qualgptbg}{RGB}{235,245,243}
\definecolor{qualpangbg}{RGB}{244,238,247}
\newlength{\qualpanelheight}
\newcommand{\qualbodyfont}{\footnotesize}
\newcommand{\qualmetric}[4]{%
  \begingroup
  \setlength{\fboxsep}{2.5pt}%
  \colorbox{#3}{\scriptsize\textsf{\textcolor{#4}{\textbf{#1:}}}~\scriptsize #2}%
  \endgroup
}
\newcommand{\qualtag}[1]{%
  \begingroup
  \setlength{\fboxsep}{2.0pt}%
  \colorbox{qualtagbg}{\scriptsize\textsf{\textcolor{qualaccent}{\textbf{#1}}}}%
  \endgroup
}
\newcommand{\qualexamplepanel}[6]{%
  \begingroup
  \setlength{\fboxsep}{7pt}%
  \fcolorbox{qualboxborder}{white}{%
    \begin{minipage}[t][\qualpanelheight][t]{0.462\linewidth}
      \raggedright
      \setlength{\parindent}{0pt}%
      {\small\bfseries\textcolor{qualaccent}{#1}\hfill\qualtag{#2}\par}
      \vspace{0.15em}
      {\color{qualaccent}\rule{\linewidth}{0.8pt}\par}
      \vspace{0.45em}
      \qualmetric{Semantic}{#3}{qualsembg}{qualaccent}\hfill\qualmetric{GPTZero}{#4}{qualgptbg}{qualaccent}\hfill\qualmetric{Pangram}{#5}{qualpangbg}{qualaccent}\par
      \vspace{0.55em}
      {\qualbodyfont #6\par}
    \end{minipage}%
  }%
  \endgroup
}

\title{Base Models Look Human To AI Detectors}

\author{%
  Yixuan Even Xu \quad Ziqian Zhong \quad Aditi Raghunathan \quad Fei Fang \quad J. Zico Kolter \\
  Carnegie Mellon University\\
  \texttt{\{yixuanx,ziqianz,aditirag,feif,zkolter\}@cs.cmu.edu} \\
}

\begin{document}

\maketitle

\begin{abstract}
As AI-generated text enters the real-world at scale, institutions increasingly use commercial AI-text detectors, especially in education and academic-integrity workflows. We report a surprising empirical finding about such systems: when evaluated by GPTZero and Pangram, generated text from base models is often judged overwhelmingly human, whereas text generated by their instruction-tuned counterparts is not. Building on this observation, we propose \textit{Humanization by Iterative Paraphrasing (HIP)}, a detector-agnostic pipeline that minimally fine-tunes a base model into a paraphraser and applies it iteratively. Compared with the baselines we test, HIP yields a stronger trade-off between semantic preservation and detector evasion on commercial detectors. Across Llama-3 and Qwen-3 families, spanning model sizes from 0.6B to 70B, HIP consistently improves detector human-likeness. Our findings suggest that current detectors are tracking artifacts of instruction tuning and local context more than any invariant notion of machine-generated text. This, in turn, calls for detector designs that model these factors more explicitly.
\end{abstract}

\section{Introduction}
\label{sec:intro}

As large language model (LLM) text becomes commonplace, distinguishing human-written text from machine-generated text has become a practical problem rather than a purely academic one. Commercial LLM-text detection systems such as GPTZero \citep{gptzero2026} and Pangram \citep{pangram2024} have emerged, and they have been deployed in real-world use cases including assignment screening and authorship review \citep{gptzerousecase,pangramusecase}.

At the same time, a growing body of work studies how to evade such detectors by treating them as optimization targets. This includes paraphrasing-based rewriting and, more recently, reinforcement-learning-based methods that optimize directly against detector APIs \citep{david2025authormist,ranganath2026stealthrl}. Our work begins one step earlier: \textit{are there models whose outputs commercial detectors already judge to be human-written, without detector-aware optimization?}

\begin{figure}[!h]
    \centering
    \includegraphics[width=\linewidth]{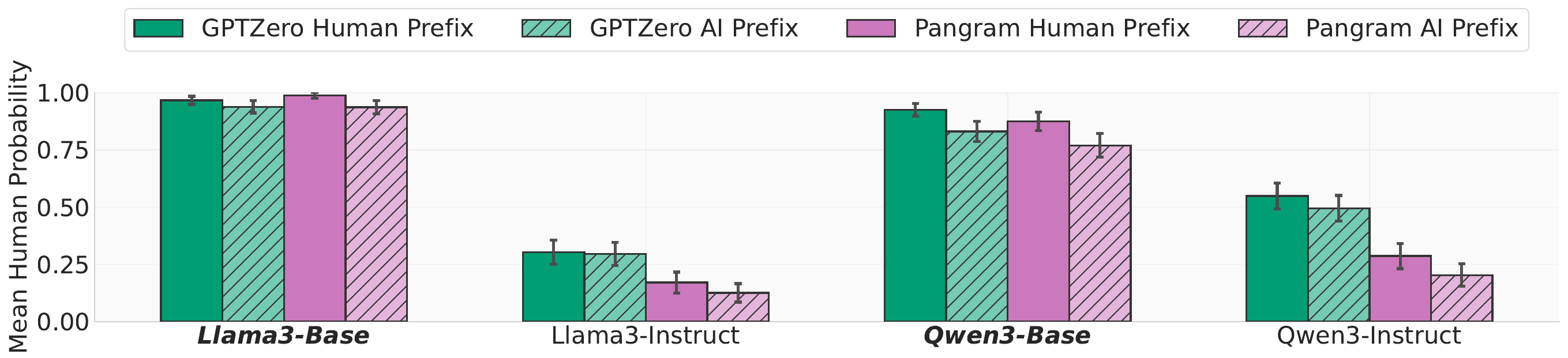}
    \vspace{-1.2em}
    \caption{GPTZero and Pangram human-probability scores on text generated by base and instruction-tuned Llama3-8B and Qwen3-8B models when conditioned on human-written or AI-generated prefixes. The error bars show $95\%$ confidence intervals. Across both model families, base-model continuations are judged substantially more human than instruct-model continuations.}
    \label{fig:intro_completion_aggregate}
    \vspace{-0.8em}
\end{figure}

The answer is yes. Current commercial detectors judge base-model continuations far more human than instruction-tuned continuations. To show this, we directly evaluate Llama-3-8B and Qwen3-8B under human-written and AI-generated single-sentence prefixes. \Cref{fig:intro_completion_aggregate} summarizes the result. For Llama-3-8B with human prefixes, GPTZero and Pangram assign human probabilities of $96.7\%$ and $98.8\%$ to the base model's continuations, respectively, and $30.3\%$ and $17.1\%$ to the instruct model's continuations. Similar gaps appear under AI prefixes and on Qwen3-8B.

These measurements suggest two working intuitions about what makes model outputs look human to current detectors. The first is \textit{low distortion}: outputs closer to base-model continuation behavior are judged more human than outputs produced after instruction tuning. The second is \textit{human context}: human prefixes make model continuations look slightly more human than AI prefixes. In other words, conditioning on text already drawn from the human-written distribution can shift subsequent continuations in a more human-looking direction from the perspective of current detectors.

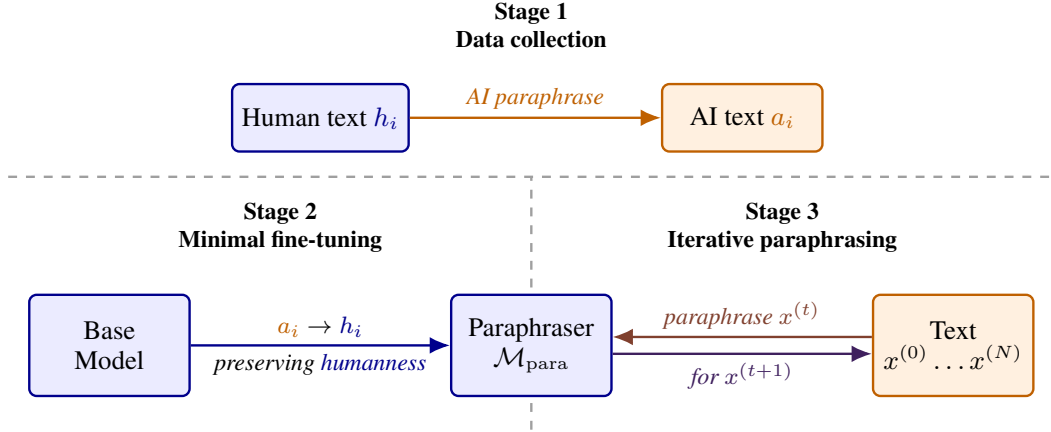
\begin{figure}[t]
    \centering
    \resizebox{\linewidth}{!}{\input{figures/method/hip_method}}
    \caption{Overview of Humanization by Iterative Paraphrasing (HIP). \textbf{Stage 1 (data collection):} an AI paraphraser rewrites each human passage $h_i$ into an AI counterpart $a_i$, yielding paired data $\mathcal{D}=\{(a_i, h_i)\}$. \textbf{Stage 2 (minimal fine-tuning):} a base model is lightly adapted on $a_i \to h_i$ to obtain a paraphraser $\mathcal{M}_{\mathrm{para}}$ while preserving native continuation behavior as much as possible. \textbf{Stage 3 (iterative paraphrasing):} $\mathcal{M}_{\mathrm{para}}$ is applied repeatedly to an input passage, producing $x^{(0)},\dots,x^{(N)}$ whose outputs become progressively more human-like to current detectors.}
    \label{fig:hip_method}
\end{figure}

The observations motivate a detector-agnostic rewriting pipeline. We minimally fine-tune a base model into a paraphraser while keeping it close to base-model continuation behavior, thereby preserving \textit{low distortion}. We then apply it iteratively so that the local context is progressively rewritten away from the original AI text and toward \textit{human context}. We call this pipeline \emph{Humanization by Iterative Paraphrasing (HIP)} and illustrate it in \cref{fig:hip_method}. Across Llama and Qwen models of multiple sizes, HIP yields a stronger trade-off between semantic retention and detector evasion on the state-of-the-art commercial detectors we study than the previous approaches we test, including simple prompt-based paraphrasing, supervised paraphrasing baselines \citep{krishna2023paraphrasing}, Unicode-substitution baselines \citep{creo2025silverspeak}, and reinforcement-learning-based detector-evasion methods \citep{ranganath2026stealthrl}. Moreover, unlike much of the academic literature, which evaluates primarily on open-source detectors, we conduct this evaluation on state-of-the-art commercial detectors.

We summarize our contributions as follows.
{\setlength{\leftmargini}{1.6em}
\begin{itemize}
    \item We identify a surprising empirical pattern on commercial detectors: base-model continuations are judged substantially more human than instruction-tuned continuations under the same prefix conditions, which motivates two intuitions about what makes model outputs look human to current detectors: \textit{low distortion} and \textit{human context}.
    \item We introduce \emph{Humanization by Iterative Paraphrasing (HIP)}, a detector-agnostic pipeline that minimally adapts a base model into a paraphraser and applies it iteratively to humanize AI-generated text. Empirically, HIP works across Llama and Qwen model families and a range of model sizes, yielding a stronger semantic-evasion trade-off than the previous approaches we test.
    \item We point to detector-side research directions, arguing that future systems should pay attention to base-model behavior, post-training distortions, and local context more explicitly.
\end{itemize}
}

\section{Related Work}
\label{sec:related}

\textbf{AI text detection.} As LLMs have advanced, detecting AI-generated text has become an important practical problem. Existing methods include zero-shot or statistical approaches, such as DetectGPT \citep{mitchell2023detectgpt} and Binoculars \citep{hans2024binoculars}, as well as supervised classifiers trained on labeled human and machine text. Commercial detectors such as Pangram \citep{pangram2024} and GPTZero \citep{gptzero2026} report strong cross-domain performance using supervised neural classifiers trained on large corpora of human- and machine-written text. As LLMs are increasingly used as collaborative co-authors rather than sole generators, the boundary between human and machine text is also blurring. \citet{editlens2025} move beyond binary classification by quantifying the extent of AI editing, while MixSet \citep{mixset} evaluates detectors in subtle revision and mixed-authorship settings. Much of this literature evaluates text produced directly by assistant-style or post-trained models. Our paper instead asks how current detectors behave on unmodified base-model continuations, especially under human-written prefix context.

\textbf{Behavior shift during post-training.} Instruction tuning and RLHF leave statistical fingerprints that can be both characterized and partially reversed. On the characterization side, \citet{casper2023open} list distributional shift as a central concern of post-training, and concrete artifacts have been documented including response length \citep{singhallong} and sycophancy \citep{sharma2023towards}. \citet{movva2026whatshumanfeedbacklearning} use sparse autoencoders to analyze preference datasets, finding that LMArena strongly favors Markdown-style formatting with headings, lists, and bolded text. On the reversibility side, \citet{jindal2025keep} document that continual pretraining significantly degrades instruction performance, and \citet{morris2025gptossbase} recover a base-like model from the post-trained GPT-OSS-20B via low-rank fine-tuning on pre-training data. Our paper contributes to both strands: we use detector behavior as an empirical lens on post-training shifts, and we find that benign continued exposure to base-style data is sufficient to recover detector human-likeness without any detector-aware optimization.

\textbf{Adversarial paraphrasing and detector evasion.} The deployment of AI text detectors has been accompanied by a growing line of research on how to evade them. \citet{sadasivan2023can} analyze paraphrasing as a fundamental weakness of many detectors, and DAMAGE \citep{damage2025} studies detectors on humanized AI text while proposing a more robust detector. Recent methods include temperature-guided paraphrasing such as TempParaphraser \citep{huang2025tempparaphraser}, supervised rewriting models such as DIPPER \citep{krishna2023paraphrasing}, orthographic attacks based on homoglyph substitution such as SilverSpeak \citep{creo2025silverspeak}, style-humanization approaches such as MASH \citep{gu2026mash}, and reinforcement-learning-based attacks such as AuthorMist \citep{david2025authormist} and StealthRL \citep{ranganath2026stealthrl}, which optimize against black-box detector APIs. Beyond the academic literature, commercial AI humanizers are also now marketed explicitly as detector-evasion tools, and recent academic work has begun to study such systems systematically \citep{damage2025}. Our paper studies detector evasion in a different regime: we use minimal adaptation to exploit a human-like behavior already present in base-model generations, evaluate on state-of-the-art commercial detectors rather than only on open or research detectors, and use the observed behavior to point toward new research directions for detectors.

\textbf{Contextual influence and iterative refinement.} The context in which an LLM operates strongly influences its generation distribution, so iterative rewriting has become a natural setting for detector evasion. TH-Bench \citep{zheng2025th} studies humanization attacks against detectors, while PADBen \citep{zha2025padben} specifically analyzes iterative paraphrasing and benchmarks robustness to paraphrase attacks. Beyond evasion, iterative refinement is also a general capability of modern LLMs. Self-Refine \citep{madaan2023self} shows that a single model can improve outputs through repeated feedback-and-revision cycles. Our paper connects these strands by asking whether iterative paraphrasing can progressively replace AI-origin context with more human-looking context.

\section{Methodology}
\label{sec:method}

We have seen in \cref{sec:intro} that base models, when conditioned on human text, are overwhelmingly detected as human by current detectors. As discussed in \cref{sec:intro}, this phenomenon suggests two central intuitions: \textit{low distortion} and \textit{human context}. HIP operationalizes these intuitions with a detector-agnostic pipeline that minimally adapts a base model into a paraphraser and then applies that paraphraser iteratively. The pipeline has three stages: data preparation, minimal fine-tuning, and iterative paraphrasing. We describe each stage in the following subsections.

\subsection{Data Preparation}
\label{subsec:method_data}

The first stage constructs paired examples
$\mathcal{D}=\{(a_i, h_i)\}_{i=1}^{M}$,
where $h_i$ is a high-quality human passage and $a_i$ is an AI paraphrase of the same passage. Here, the direction of the pair matters: we will ultimately train a model to map from the AI text back to the human text.

As summarized in \cref{alg:data_prep}, the raw corpus is first narrowed to a candidate set by applying basic corpus filters, for example on provenance, length, or document integrity. These candidates are then normalized into a common textual form and deduplicated at the corpus level. After that, a text-quality screen removes passages that are poor targets for rewriting. Only then do we construct pairs. For each remaining human passage $h_i$, an external paraphraser generates an AI-style rewrite $a_i$. Pair construction uses bounded rejection and re-sampling: candidates that fail anomaly checks or semantic-preservation checks are discarded and regenerated, and the example is dropped if no valid paraphrase is obtained within a fixed retry budget. Essentially, HIP constructs and trains on \emph{filtered human targets} and \emph{meaning-preserving AI-style sources}, rather than on arbitrary raw text.

\begin{algorithm}[h]
\caption{Data Preparation}
\label{alg:data_prep}
\begin{algorithmic}[1]
\Require A human corpus $\mathcal{C}_{\mathrm{raw}}$ and the paraphrasing retry budget $K$
\Ensure Paired dataset $\mathcal{D}=\{(a_i, h_i)\}$
\State $\mathcal{C}_{\mathrm{cand}} \gets \textsc{FilterByProvenanceAndLength}(\mathcal{C}_{\mathrm{raw}})$
\State $\mathcal{C}_{\mathrm{dedup}} \gets \textsc{Deduplicate}(\textsc{Normalize}(\mathcal{C}_{\mathrm{cand}}))$
\State $\mathcal{C}_{\mathrm{clean}} \gets \{h \in \mathcal{C}_{\mathrm{dedup}} : \textsc{TextQualityOK}(h)\}$
\State $\mathcal{D} \gets \varnothing$
\For{passage $h \in \mathcal{C}_{\mathrm{clean}}$}
    \For{attempt $\in \{1, \dots, K\}$}
        \State $a \gets \textsc{AIParaphrase}(h)$
        \If{\textsc{AnomalyFree}($a$) and \textsc{SemanticPreservationOK}($a, h$)}
            \State $\mathcal{D} \gets \mathcal{D} \cup \{(a, h)\}$
            \State \textbf{break}
        \EndIf
    \EndFor
\EndFor
\State \Return $\mathcal{D}$
\end{algorithmic}
\end{algorithm}

\subsection{Minimal Fine-Tuning}
\label{subsec:method_ft}

Given the paired dataset $\mathcal{D}$, the second stage trains a paraphraser $\mathcal{M}_{\mathrm{para}}$ from a pretrained language model $\mathcal{M}_{\mathrm{base}}$ while perturbing the model as little as possible to preserve \textit{low distortion}. HIP therefore uses \emph{minimal fine-tuning}: we do not train a full assistant. Instead, we apply supervised fine-tuning to $\mathcal{D}$, optionally with a parameter-efficient update such as low-rank adaptation \citep{hu2022lora}.

The supervision format is likewise kept simple. Rather than using a chat template, we consider paraphrasing as a plain text continuation problem with lightweight structural tags. For a single pair $(a, h)$, where $a$ is the AI paraphrase and $h$ is the original human passage, the model sees:

\begin{lstlisting}[basicstyle=\ttfamily\small, frame=single, framerule=0.3pt, rulecolor=\color{black!25}, backgroundcolor=\color{black!2}, xleftmargin=1em, xrightmargin=1em, aboveskip=0.75em, belowskip=0.75em]
<source_text>
  AI paraphrase a
</source_text>

<target_text>
  Original human passage h
</target_text>
\end{lstlisting}

Operationally, the text between the \texttt{<source\_text>} tag and the \texttt{<target\_text>} tag form the prompt prefix, while the original human passage and closing \texttt{</target\_text>} tag form the completion. Training then uses the standard next-token objective, but the loss is restricted to the completion span only. In other words, the model is optimized to reconstruct the human passage conditioned on the AI paraphrase, not to imitate a conversational interface via a chat template.

\subsection{Iterative Paraphrasing}
\label{subsec:method_inference}

Once the paraphraser is trained, the final stage applies it to transform a machine-like passage $x^{(0)}$ into a rewrite $x^{(N)}$ through \textit{iterative paraphrasing}. The use of iteration is deliberate: a single pass may still retain residual features of the original text, whereas multiple rounds progressively build \textit{human context}. We therefore apply the paraphraser for a fixed number of rounds, producing a sequence $x^{(0)}, x^{(1)}, \dots, x^{(N)}$, where each $x^{(t)}$ rewrites the previous round's output.

In execution, \cref{alg:iterative} reuses the same prompt structure as training at every round. The current passage $x^{(t-1)}$ is placed into the source field, and the model generates a new target passage $x^{(t)}$. As the number of rounds increases, the semantic content of the text may gradually drift, but the text also moves away from the statistical region occupied by the original generator and toward the paraphraser's own preferred continuation regime. This trades off semantic retention for humanization.

\begin{algorithm}[h]
\caption{Iterative Paraphrasing}
\label{alg:iterative}
\begin{algorithmic}[1]
\Require The initial text $x^{(0)}$, the paraphraser $\mathcal{M}_{\mathrm{para}}$, and the number of iterations $N$
\Ensure The final paraphrased text $x^{(N)}$
\For{$t \in \{1, \dots, N\}$}
    \State $p^{(t)}$ $\gets$ \textsc{FormatAsPrompt}($x^{(t-1)}$)
    \State $x^{(t)} \gets \mathcal{M}_{\mathrm{para}}(p^{(t)})$
\EndFor
\State \Return $x^{(N)}$
\end{algorithmic}
\end{algorithm}

\section{Experiments}
\label{sec:experiments}

\begin{figure*}[p]
    \centering
    \includegraphics[width=\textwidth]{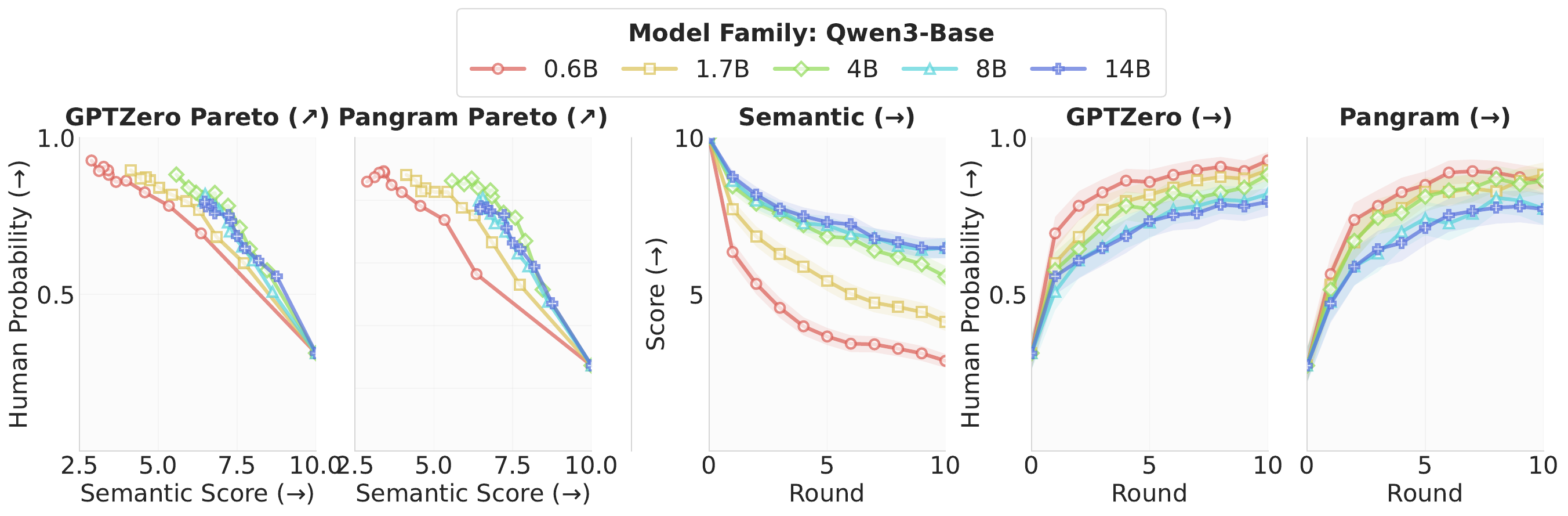}

    \vspace{0.75em}

    \includegraphics[width=\textwidth]{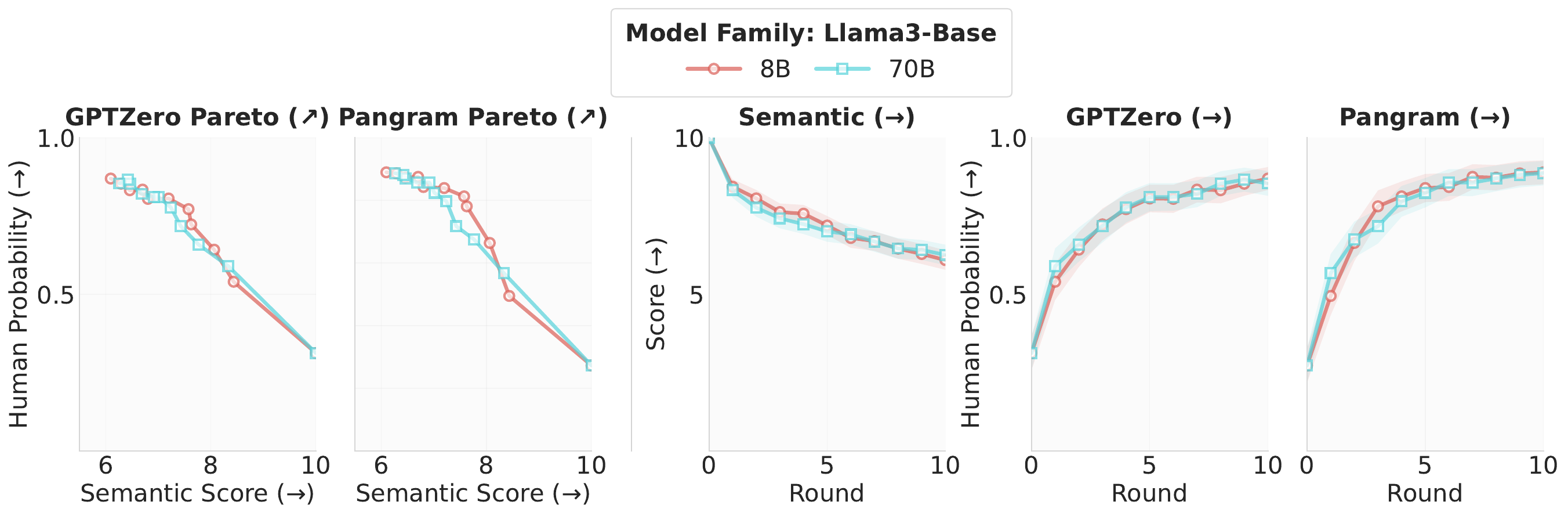}

    \vspace{0.75em}

    \includegraphics[width=\textwidth]{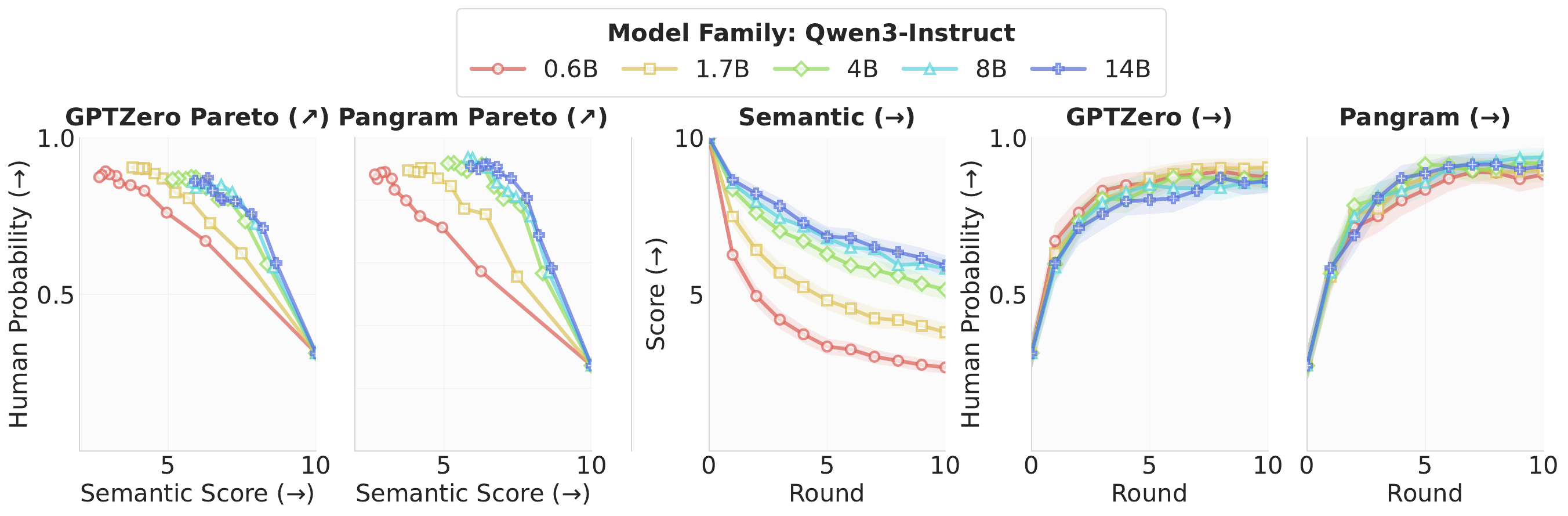}

    \vspace{0.75em}

    \includegraphics[width=\textwidth]{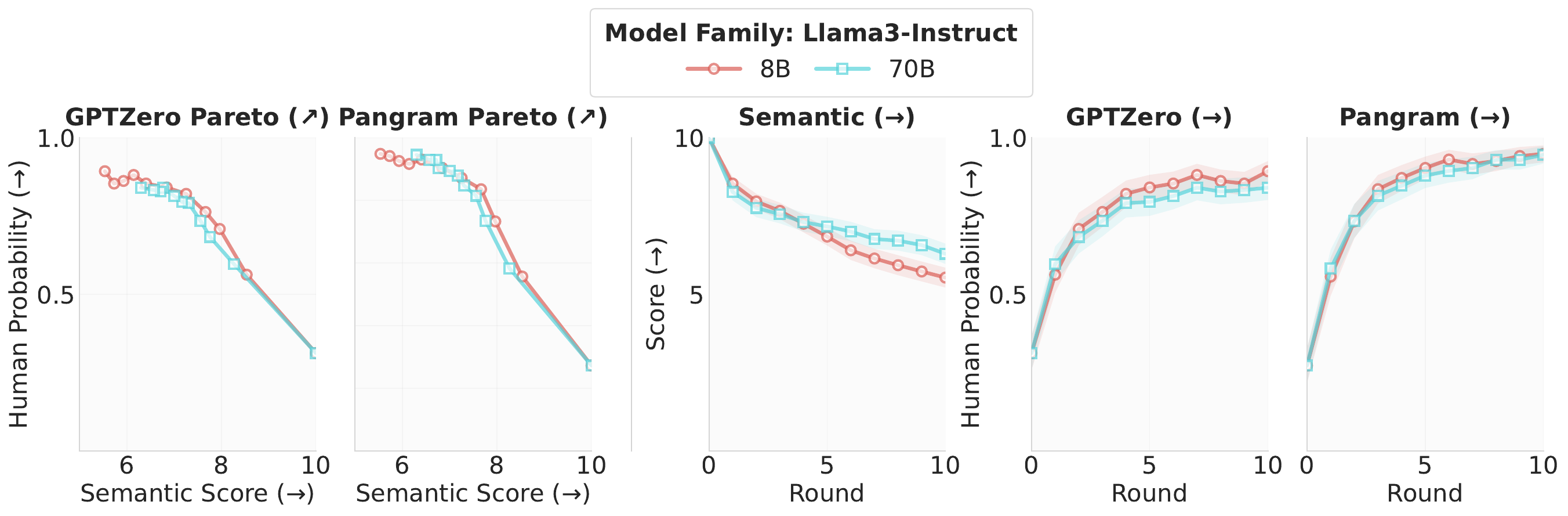}
    \caption{HIP across model families. From top to bottom: Qwen3-Base, Llama3-Base, Qwen3-Instruct, and Llama3-Instruct. Within each panel, the first two subplots show the GPTZero and Pangram Pareto frontiers for the trade-off between semantic preservation and human-likeness, while the last three show semantic score, GPTZero human probability, and Pangram human probability over the rounds, respectively. Shaded areas indicate $95\%$ confidence intervals.}
    \label{fig:model_families}
\end{figure*}

\begin{figure}[ht]
    \centering
    \includegraphics[width=\linewidth]{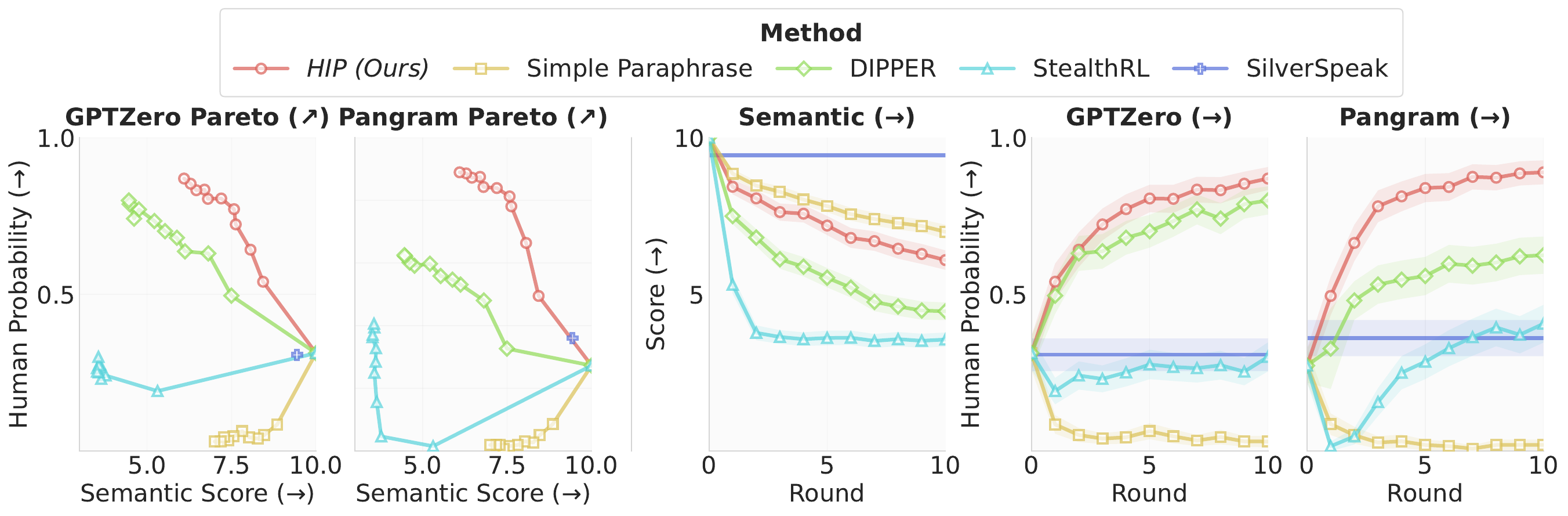}
    \caption{HIP versus baseline methods. The first two subplots show the GPTZero and Pangram Pareto frontiers for the trade-off between semantic preservation and human-likeness, while the last three show semantic score, GPTZero human probability, and Pangram human probability over the rounds, respectively. Shaded areas indicate $95\%$ confidence intervals. HIP traces the strongest trade-off between semantic preservation and human-likeness among the methods we tested.}
    \label{fig:hip_baselines}
\end{figure}

In this section, we evaluate HIP as a paraphrase-based detector-evasion method across model families, sizes, and baseline methods. We also describe the continuation evaluation introduced in \cref{sec:intro}. We release our code for training and running HIP at \url{https://github.com/YixuanEvenXu/humanization-by-iterative-paraphrasing}. We release the training and evaluation data, together with the LoRA adapters, through the Hugging Face collection at \url{https://huggingface.co/collections/YixuanEvenXu/humanization-by-iterative-paraphrasing}.

\subsection{Experimental Setup}

\paragraph{Datasets.} Our experiments require both human-written texts and AI-generated texts from the same domains. We use selected subsets of RAID \citep{dugan2024raid} and MAGE \citep{li2024mage}, targeting clean, document-style prose. From RAID, we keep the domains \texttt{abstracts}, \texttt{books}, \texttt{news}, and \texttt{wiki}. From MAGE, we keep the human source families \texttt{xsum\_human}, \texttt{cnn\_human}, \texttt{tldr\_human}, and \texttt{squad\_human}, together with their AI counterparts \texttt{xsum}, \texttt{cnn}, \texttt{tldr}, and \texttt{squad}.
{\setlength{\leftmargini}{1.6em}
\begin{itemize}
    \item \textbf{Training set.} We construct the paired dataset as described in \cref{alg:data_prep}, with $\mathcal{C}_{\mathrm{raw}}$ being the selected human corpus. After filtering, deduplication, and text-quality screening, each remaining human passage $h_i$ is paraphrased by GPT-5-nano into $a_i$ to form the supervised dataset $\mathcal{D}=\{(a_i,h_i)\}_{i=1}^{M}$. This process yields a dataset of $11757$ training pairs.
    \item \textbf{Evaluation set.} The main evaluation set consists of $256$ AI-generated passages, constructed by taking the first $32$ examples from each of the eight retained RAID and MAGE source categories.
\end{itemize}}

\paragraph{Evaluation metrics.} When evaluating a paraphrasing model or method on the evaluation set, we report three primary metrics. The first is semantic preservation, scored by GPT-5-nano on an integer scale from $0$ to $10$ by comparing each rewritten text against its original input. A score of $10$ denotes complete preservation of meaning, while lower scores indicate greater semantic drift. The other two are detector-based human-likeness scores from the commercial systems GPTZero \citep{gptzero2026} and Pangram \citep{pangram2024}. Both detectors return probability distributions over authorship labels, and we report the probability assigned to the \textit{human} label. Higher values on both detector metrics therefore indicate that a rewritten text is judged more human-like. For qualitative examples that illustrate how these metrics align with actual outputs, see \cref{app:qual_examples}.

\paragraph{Models.} We conduct experiments on both base and instruction-tuned models from the Qwen3 family \citep{yang2025qwen3} and the Llama3 family \citep{grattafiori2024llama}. For Qwen3, we use the 0.6B, 1.7B, 4B, 8B, and 14B models. For Llama3, we use the 8B and 70B models.

\paragraph{Fine-tuning and inference configurations.} For each selected model, whether base or instruction-tuned, we apply the same minimal fine-tuning procedure on the training set $\mathcal{D}$ to obtain a paraphraser $\mathcal{M}_{\mathrm{para}}$, using the plain source-target format from \cref{subsec:method_ft}. All runs use one epoch of training, a maximum sequence length of $2048$, effective batch size $16$, learning rate $5\times 10^{-5}$ with cosine scheduling, and LoRA \citep{hu2022lora} with rank $128$, scaling factor $128$, and dropout $0.05$. For the 70B models, training uses QLoRA \citep{dettmers2023qlora} for memory efficiency. Inference for all models is served with vLLM \citep{kwon2023efficient}. At inference time, we apply $\mathcal{M}_{\mathrm{para}}$ iteratively for $N=10$ rounds. Across all runs, generation uses temperature $1.0$ and top-$p$ $0.95$.

\paragraph{Baseline methods.} We compare HIP against several representative detector-evasion baselines. The set of possible baselines is large and growing, and we do not aim to exhaust it. Instead, we choose the set of baselines that span different types of approaches and have released checkpoints:
{\setlength{\leftmargini}{1.6em}
\begin{itemize}
    \item \textit{Simple Paraphrase}: Directly applying a zero-shot paraphrase prompt at inference time.
    \item \textit{DIPPER} \citep{krishna2023paraphrasing}: A supervised paraphrasing method that aims to preserve meaning while varying surface form, using lexical and sentence-level controls to steer diversity.
    \item \textit{SilverSpeak} \citep{creo2025silverspeak}: A Unicode homoglyph-substitution method that perturbs token appearance without rewriting the text, targeting detector sensitivity to character-level cues.
    \item \textit{StealthRL} \citep{ranganath2026stealthrl}: A reinforcement-learning-based detector evasion method that optimizes a paraphraser against open-source detectors.
\end{itemize}}

\paragraph{Continuation evaluation.} For the continuation evaluation introduced in \cref{sec:intro}, we use $256$ human-written and $256$ AI-generated passages from the same selected RAID and MAGE domains. The prefixes are truncated to their first sentence and then used as continuation prompts. For each prefix, we generate one continuation and score only the generated text for human-likeness with GPTZero and Pangram. This evaluation compares the human-likeness of continuations from base and instruction-tuned models from the Qwen3 and Llama3 families as shown in \cref{fig:intro_completion_aggregate}. In \cref{app:continuation}, we extend this setup to include HIP-adapted and continued-pretraining controls.

\paragraph{Computation and API cost.} The experiments were conducted on GPU nodes with either $4$ or $8$ NVIDIA L40S GPUs. In total, the local training and inference runs consume roughly $300$ GPU-hours. Across the project, OpenAI API usage for dataset construction, semantic scoring, and model fine-tuning cost roughly $200$ dollars. At list prices, commercial-detector evaluation would have been more expensive: our GPTZero usage totaled about $40$ million words, costing about $6000$ dollars, and our Pangram usage totaled about $100000$ passages, costing about $5000$ dollars. GPTZero and Pangram provided research access to their models. To our knowledge, relatively few papers report detector-evasion results on state-of-the-art commercial detectors rather than only on open-source detectors, which strengthens the empirical relevance of our evaluation.

\subsection{Experimental Results}

\begin{figure*}[t]
    \centering
    \setlength{\qualpanelheight}{2.70in}
    \qualexamplepanel{Original AI Text}{MAGE / XSum}{$10$ (ref.)}{$0.0$\%}{$0.0$\%}{A police force has apologised after its officers made a sandcastle ``crime scene'' featuring a naked dead woman. The incident happened during a training exercise on Bournemouth beach in Dorset, England. The force said the scene was ``clearly inappropriate'' and an investigation has been launched. It is not yet clear how the scene came to be created or who was responsible. The incident has caused outrage on social media, with many people calling for the officers involved to be disciplined. This is not the first time that the police have been accused of being insensitive in their training exercises. In 2016, the Metropolitan Police in London was criticised for using photos of real victims of knife crime in a training session.}\hfill
    \qualexamplepanel{HIP (Llama3-8B) Round 1}{MAGE / XSum}{$8$}{$8.2$\%}{$0.0$\%}{A police force has apologised after its officers created a ``crime scene'' sandcastle, including a naked dead woman. Dorset Police said officers from its rural crime unit were taking part in a training exercise on Bournemouth beach in the south west of England. The force said the scene was ``clearly inappropriate'' and an investigation was under way. It is not yet known how the sandcastle was created or who was responsible. The incident has been condemned on social media and people have called for the officers involved to be disciplined. It is not the first time police have been accused of being insensitive in their training. In 2016, the Metropolitan Police in London faced criticism for using photos of actual victims of knife crime in a training exercise.}

    \vspace{0.8em}

    \qualexamplepanel{HIP (Llama3-8B) Round 2}{MAGE / XSum}{$9$}{$20.9$\%}{$100.0$\%}{A police force has apologised for the actions of officers who created a ``crime scene'' sandcastle including a naked dead woman. Dorset Police said officers from its rural crime unit were taking part in a training exercise on Bournemouth beach in the south west of England. The force said it was an ``inappropriate'' scene and was investigating. It is not known how the sandcastle was made and who was responsible. The incident has been criticised on social media, with people calling for the officers involved to be disciplined. It is not the first time police have been criticised for appearing insensitive in their training. In 2016, the Metropolitan Police in London was criticised for using photographs of victims of knife crime in their training.}\hfill
    \qualexamplepanel{HIP (Llama3-8B) Round 10}{MAGE / XSum}{$9$}{$100.0$\%}{$100.0$\%}{Dorset Police has apologised after officers were filmed building a sandcastle. It included the words crime scene and a naked dead woman. Dorset Police said it was ``appalled'' officers were caught on camera building the sandcastle on a rural crime team training day. The officers were at Bournemouth beach, in the south of England. Dorset Police said it was investigating the ``inappropriate scene.'' It was not clear how the sandcastle was constructed or how many officers were involved. Officers were criticised on social media, with many saying those involved should be disciplined. Police have been accused of being insensitive in the way they treat victims of crime and their families. The Met police in London was criticised heavily in November 2018 for using images of victims of knife crime in their training.}
    \caption{One qualitative Llama3-8B HIP trajectory from the main evaluation set, using an AI-generated source text from the \texttt{xsum} subset of the MAGE dataset. Each panel shows the text at a different round together with its semantic score and detector-assigned human probability from GPTZero and Pangram. This qualitative example illustrates how HIP can preserve the core content while progressively shifting detector judgments toward human.}
    \label{fig:qual_example_main}
\end{figure*}

\paragraph{HIP humanizes AI-generated text across model families and scales.} 
We show in \cref{fig:model_families} the results of applying HIP to base and instruct checkpoints from the Qwen3 and Llama3 families. Each subplot represents one model family, and each line represents one model size. Within each subplot, the first two panels show the GPTZero and Pangram Pareto frontiers for the trade-off between semantic preservation and detector evasion, while the last three show how semantic score and detector-specific human probabilities change over iterative rounds of paraphrasing.

The main pattern is consistent across all four families. After training with HIP, as the paraphraser is applied for more rounds, detector-assigned human probability rises on both GPTZero and Pangram, while semantic fidelity gradually declines. In other words, the method works by moving model outputs toward a more human-like region of the detector space, but it does so at a semantic cost. This trend holds for both base and instruction-tuned checkpoints, which indicates that the humanization effect of HIP is not specific to a single model family, size, or post-training state.

Model size primarily affects the trade-off when the size is low, rather than improving it uniformly. Within Qwen3, moving from the smaller checkpoints to 4B materially improves the trade-off curve, but beyond 4B the frontiers shift only modestly and not necessarily for the better. Llama3 shows the same qualitative pattern: the 70B models are slightly more semantically stable than the 8B models, but both achieve similar trade-offs. Our interpretation is that HIP mainly requires a model large enough to paraphrase competently. Once that threshold is reached, the method largely works.

\paragraph{Qualitative examples.} Aggregate trade-off curves correspond to recognizable local edits at the example level. \Cref{fig:qual_example_main} shows one Llama3-8B HIP trajectory from the main evaluation set. Across rounds, the model preserves the core factual content while progressively rewriting phrasing and local structure. In this example, both detectors initially flag the source text as AI-generated. By round 2, GPTZero has moved into a more ambiguous regime and Pangram has already flipped to human. By round 10, semantic fidelity remains high while both detectors score the output as human. In \cref{app:qual_examples}, we present additional qualitative examples from HIP as well as the baseline methods.

\paragraph{Baseline comparison.} We compare HIP against all baselines in \cref{fig:hip_baselines}. For every model-based baseline, we evaluate the released checkpoint from the original paper, and Simple Paraphrase is the only prompting-based baseline. As shown in \cref{fig:hip_baselines}, under the commercial detectors we study, only DIPPER still achieves a non-trivial semantic-evasion trade-off among the methods we tested. SilverSpeak raises Pangram human probability slightly, but has little effect on GPTZero. Both Simple Paraphrase and StealthRL fail to achieve a meaningful trade-off on either detector. We therefore use Llama3-8B as HIP's comparison point, since it is smaller than DIPPER's 11B backbone and thus provides a conservative comparison. As \cref{fig:hip_baselines} shows, HIP on Llama3-8B substantially outperforms all tested baselines, achieving the strongest trade-off frontier under both detectors.

\paragraph{Ablation studies and additional experiments.} We defer additional experiments to \cref{app:ablation_additional}, including details about continuation evaluation (\cref{app:continuation}), HIP on OpenAI models via the fine-tuning API (\cref{app:openai_models}), running HIP on instruct models with native chat templates (\cref{app:chat_template}), and output-layer-only adaptation (\cref{app:last_layer}).

\section{Discussion}
\label{sec:discussion}

\paragraph{Why HIP works.} Our results suggest that detector-judged human-likeness improves when the model is trained directly on human tokens. Base models have this property because pretraining is almost entirely human text. HIP appears to recover part of the same behavior, but now under a paraphrase objective. The model is updated on paired rewrites whose targets are human-written passages, and iterative application reinforces that effect by progressively removing AI-generated context from the passage. On this view, HIP works not because it discovers a special decoding trick, but because it reintroduces a human-text training signal while enabling iterative paraphrasing.

\paragraph{Implications for detectors.} Our findings suggest that current detectors are not primarily detecting ``AI'' in general. Rather, they appear to be sensitive to the statistical effects of post-training and to the context in which a passage is generated. This motivates two research questions for detector design:
\begingroup
\setlength{\leftmargini}{1.6em}
\begin{enumerate}
    \item Does post-training inherently leave detector-visible statistical fingerprints, or are current alignment pipelines simply introducing artifacts that better training procedures could avoid?
    \item Can current statistical or neural detector paradigms reliably identify text from true base models, rather than primarily identifying post-training distortions?
\end{enumerate}
\endgroup

\paragraph{Limitation and broader impact.} A main limitation of this work is that detector-evasion results are unlikely to remain static. Commercial detectors can adapt in response to known attacks, so a method that works today may weaken as detectors are updated. The broader impact is mixed. On the one hand, HIP could be misused to evade deployed detectors. On the other hand, that risk already exists in practice, and the paper makes it more explicit by identifying a concrete failure mode on current commercial systems. We view the main contribution as diagnostic: identifying this weakness should help detector researchers build stronger methods against paraphrase-based evasion.

\section*{Acknowledgments}
This work was supported in part by NSF grant IIS-2046640 (CAREER).

This work was also supported by GPTZero and Pangram's research credits.

\newpage
\bibliographystyle{unsrtnat}
\bibliography{ref}

\newpage
\appendix

\section{Ablation Studies and Additional Experiments}
\label{app:ablation_additional}

\subsection{Continuation Evaluation}
\label{app:continuation}

Here we extend the continuation evaluation in \cref{fig:intro_completion_aggregate} to include two additional instruct-model-based controls: \textit{Instruct + HIP} and \textit{Instruct + FT}. The setup is otherwise the same as in \cref{sec:experiments}: we prompt each model with the same human and AI prefixes and score only the generated continuation. \textit{Instruct + HIP} denotes an instruction-tuned model after HIP adaptation. In this experiment, however, we do not use that checkpoint as a paraphraser. Instead, we use it simply as a generator, exactly as we do for the raw base and raw instruct models. \textit{Instruct + FT} denotes a separate control in which we further fine-tune the same instruct model on FineWeb-Edu \citep{penedo2024fineweb} with a plain language-modeling objective rather than the HIP paraphrase objective.

\begin{figure}[h]
    \centering
    \includegraphics[width=\linewidth]{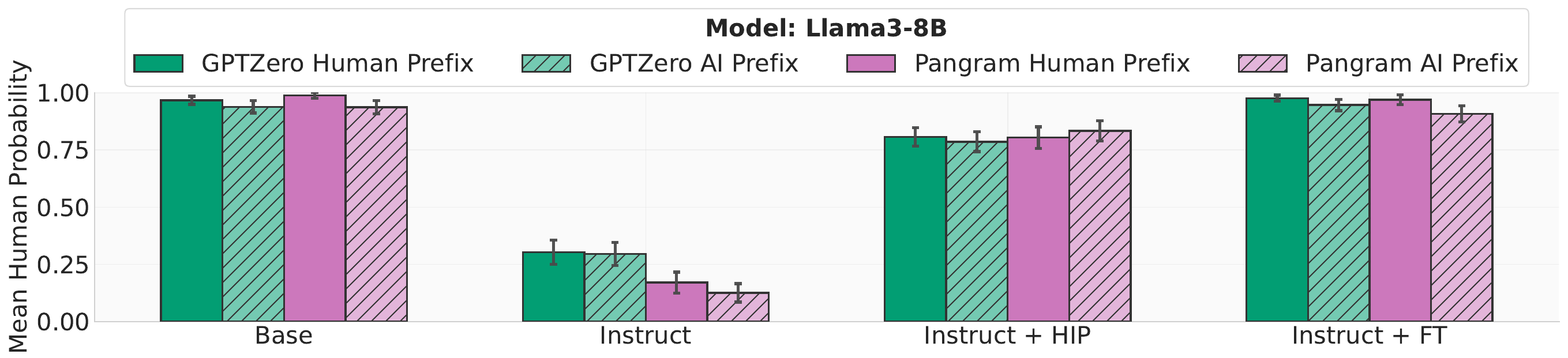}

    \vspace{0.75em}

    \includegraphics[width=\linewidth]{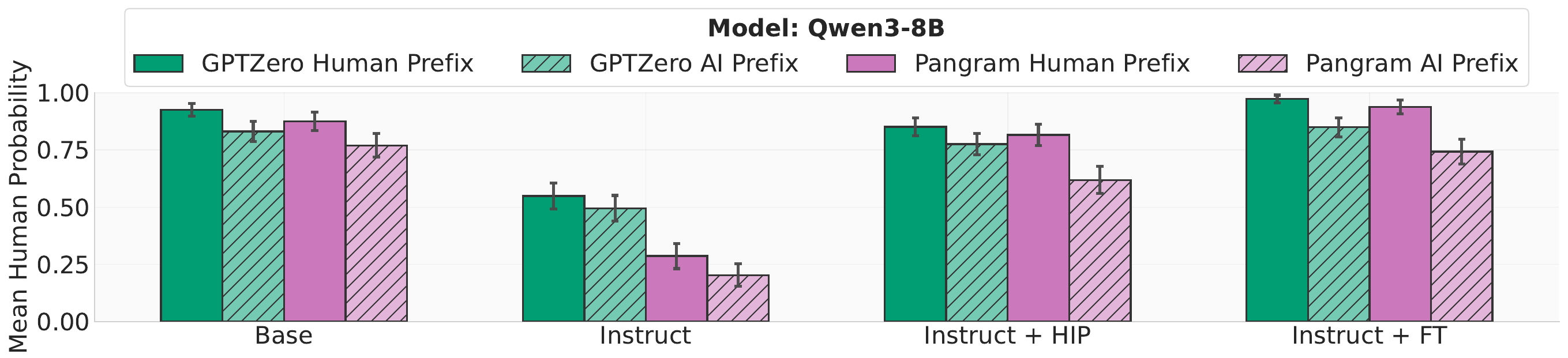}
    \caption{GPTZero and Pangram human-probability scores on text generated by Llama3-8B and Qwen3-8B variants when conditioned on human-written or AI-generated prefixes. The error bars show $95\%$ confidence intervals. In both families, the base model is judged much more human than the instruct model, and both \textit{Instruct + HIP} and \textit{Instruct + FT} recover much of that lost human-likeness.}
    \label{fig:app_continuation}
\end{figure}

As \cref{fig:app_continuation} shows, the same qualitative pattern holds in both model families and under both human and AI prefixes: instruction tuning makes continuations look much less human to GPTZero and Pangram, while both \textit{Instruct + HIP} and \textit{Instruct + FT} recover a large portion of that lost human-likeness. This suggests that detector judgments are sensitive to post-training state itself, not only to the explicit paraphrase pipeline. At the same time, \textit{Instruct + FT} does not produce a usable paraphraser, so recovering human-like continuation behavior alone is not sufficient for the main HIP objective.

\newpage
\subsection{HIP on OpenAI Models via the Fine-Tuning API}
\label{app:openai_models}

We also test whether HIP transfers to a closed-weight model fine-tuned through the OpenAI API. Starting from GPT-4.1-nano, we fine-tune the model on the same paired paraphrase dataset $\mathcal{D}$ used in the main experiments, but now through the OpenAI fine-tuning API rather than local LoRA-based training. We then run the same $10$-round iterative paraphrasing evaluation on the same $256$-example evaluation set and score the outputs with the same semantic, GPTZero, and Pangram metrics.

\begin{figure}[h]
    \centering
    \includegraphics[width=\linewidth]{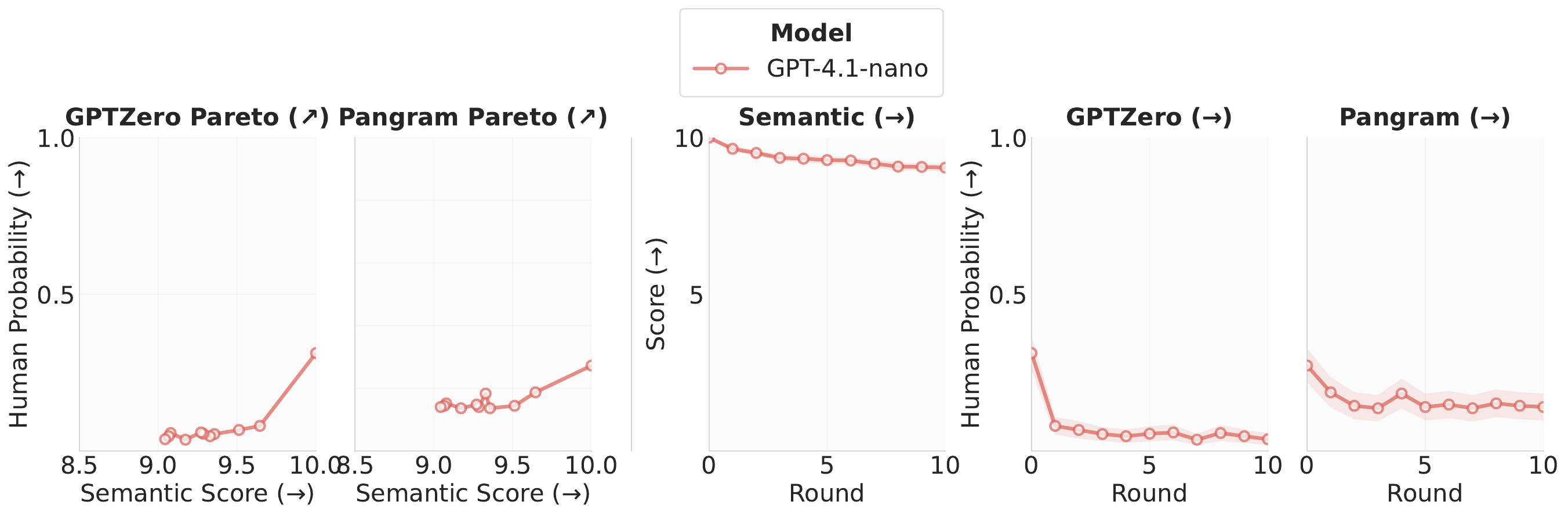}
    \caption{HIP on OpenAI GPT-4.1-nano by running through the OpenAI fine-tuning API. The first two subplots show the GPTZero and Pangram Pareto frontiers for the trade-off between semantic preservation and human-likeness, while the last three show semantic score, GPTZero human probability, and Pangram human probability over the rounds, respectively. Shaded areas indicate $95\%$ confidence intervals. Unlike the open-weight Llama and Qwen models in the main paper, the trade-off remains weak under both detectors despite relatively high semantic preservation.}
    \label{fig:app_openai}
\end{figure}

As \cref{fig:app_openai} shows, HIP does not produce a useful semantic-evasion trade-off in this setting. Semantic scores remain high throughout the $10$ rounds, but detector-assigned human probability quickly falls and remains low on both GPTZero and Pangram. In other words, the model continues to paraphrase coherently, but the iterative pipeline does not move its outputs toward the human-like region that HIP reaches on open-weight Llama and Qwen models in \cref{sec:experiments}.

Combined with the main results and the continuation evaluation, one plausible hypothesis is that fine-tuning through the OpenAI API does not induce the same low-distortion base-model behavior that HIP exploits in open-weight models. A likely reason is that the platform may mix in additional post-training or alignment-related data or procedures during training and serving. We cannot verify that mechanism directly, but the negative result is consistent with that interpretation.

\newpage
\subsection{HIP on Instruct Models with Native Chat Templates}
\label{app:chat_template}

We also test whether the humanizing effect of HIP depends on the simplified source-target formatting used in \cref{subsec:method_ft} rather than on an instruct model's native chat interface. Starting from the Llama3-8B-Instruct and Qwen3-8B-Instruct checkpoints, we keep the same paired paraphrase dataset $\mathcal{D}$ and $10$-round evaluation, but express both training and inference through each model's native chat template. Concretely, the system prompt instructs the model to paraphrase naturally while preserving meaning, and the user message contains the passage to be rewritten. We then evaluate the resulting chat-template paraphrasers on the same $256$-example evaluation set with the same metrics.

\begin{figure}[h]
    \centering
    \includegraphics[width=\linewidth]{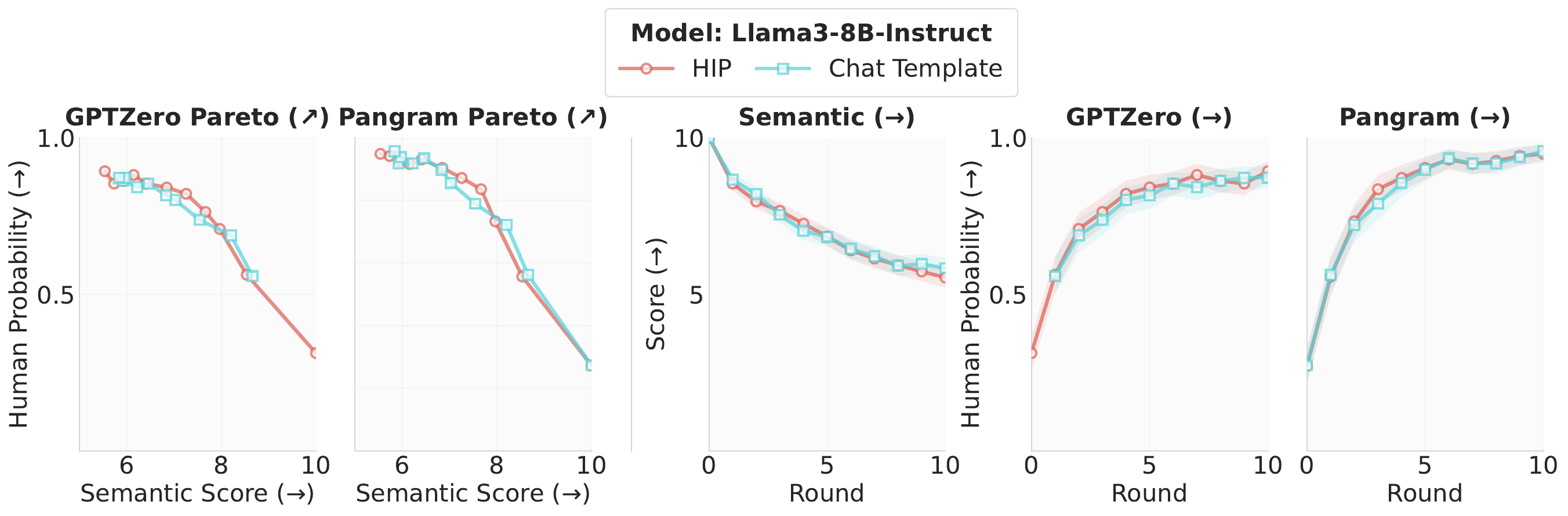}

    \vspace{0.75em}

    \includegraphics[width=\linewidth]{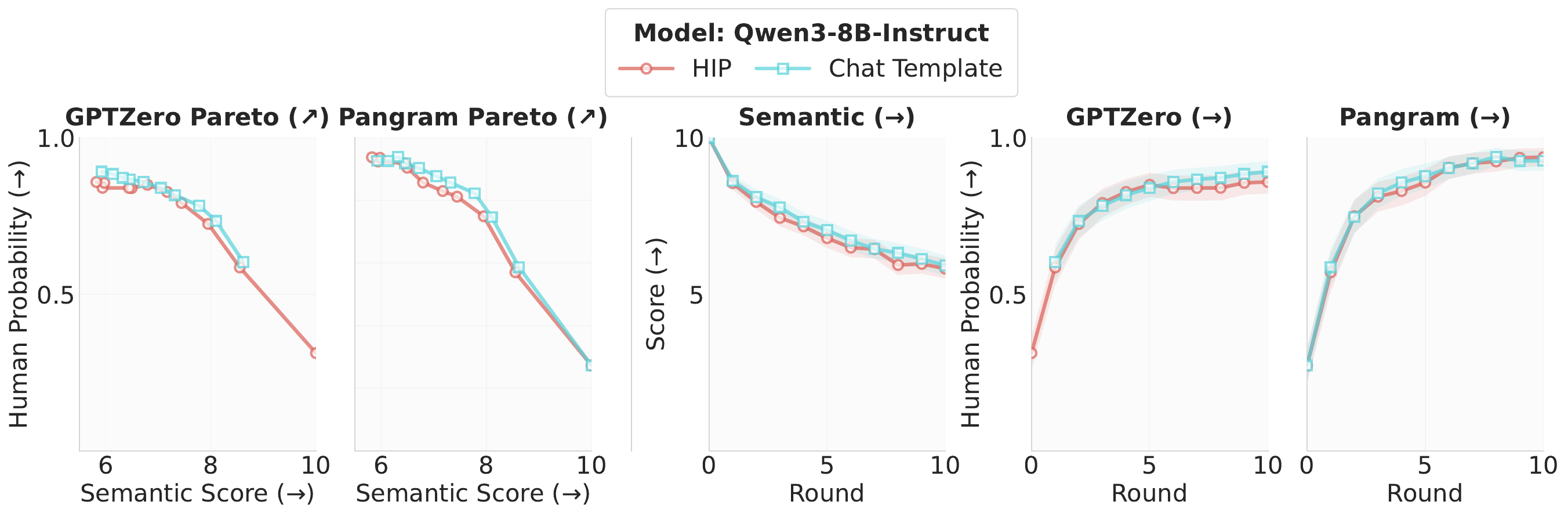}
    \caption{Standard HIP versus the native-chat-template variant on Llama3-8B-Instruct and Qwen3-8B-Instruct. In each panel, the first two subplots show the GPTZero and Pangram Pareto frontiers for the trade-off between semantic preservation and human-likeness, while the last three show semantic score, GPTZero human probability, and Pangram human probability over the rounds, respectively. Shaded areas indicate $95\%$ confidence intervals. The differences are modest in both families, suggesting that the main HIP effect is not driven by the absence of a chat template when training an instruct model.}
    \label{fig:app_chat_template}
\end{figure}

As \cref{fig:app_chat_template} shows, this change has only a modest effect on the trade-off. In both cases, the overall Pareto frontier remains close to that of standard HIP, and the qualitative behavior is unchanged.

Taken together, these results suggest that the main HIP effect does not depend strongly on whether instruct models are adapted and run through plain source-target formatting or through their native chat templates. The non-chat formulation used in \cref{subsec:method_ft} is not necessary for instruct models.

\newpage
\subsection{HIP with Output-Layer-Only Adaptation}
\label{app:last_layer}

We also test whether HIP can be approximated by changing only the final mapping from hidden states to logits. Starting from the Llama3-8B and Qwen3-8B base checkpoints, we keep the same paired paraphrase dataset $\mathcal{D}$ and the same $10$-round evaluation protocol, but freeze the entire model except for the output layer $\texttt{lm\_head}$. This ablation therefore trains a single output projection rather than the standard LoRA-based paraphraser used in the main paper. We then evaluate the resulting models on the same $256$-example evaluation set with the same semantic, GPTZero, and Pangram metrics.

\begin{figure}[h]
    \centering
    \includegraphics[width=\linewidth]{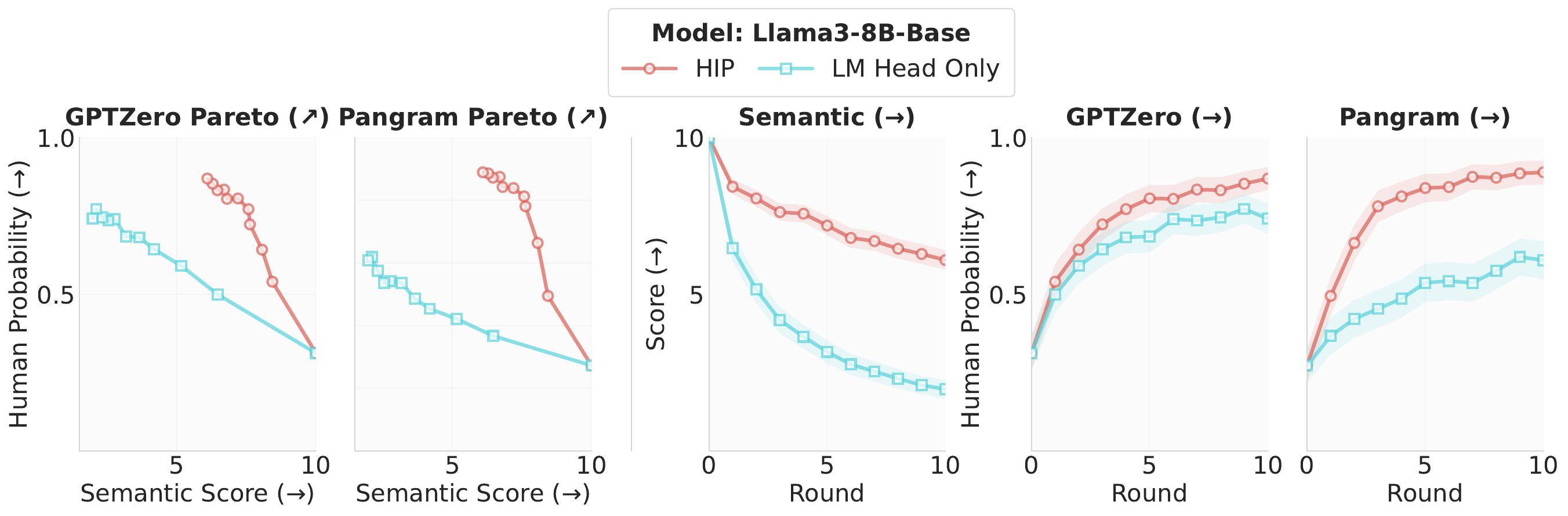}

    \vspace{0.75em}

    \includegraphics[width=\linewidth]{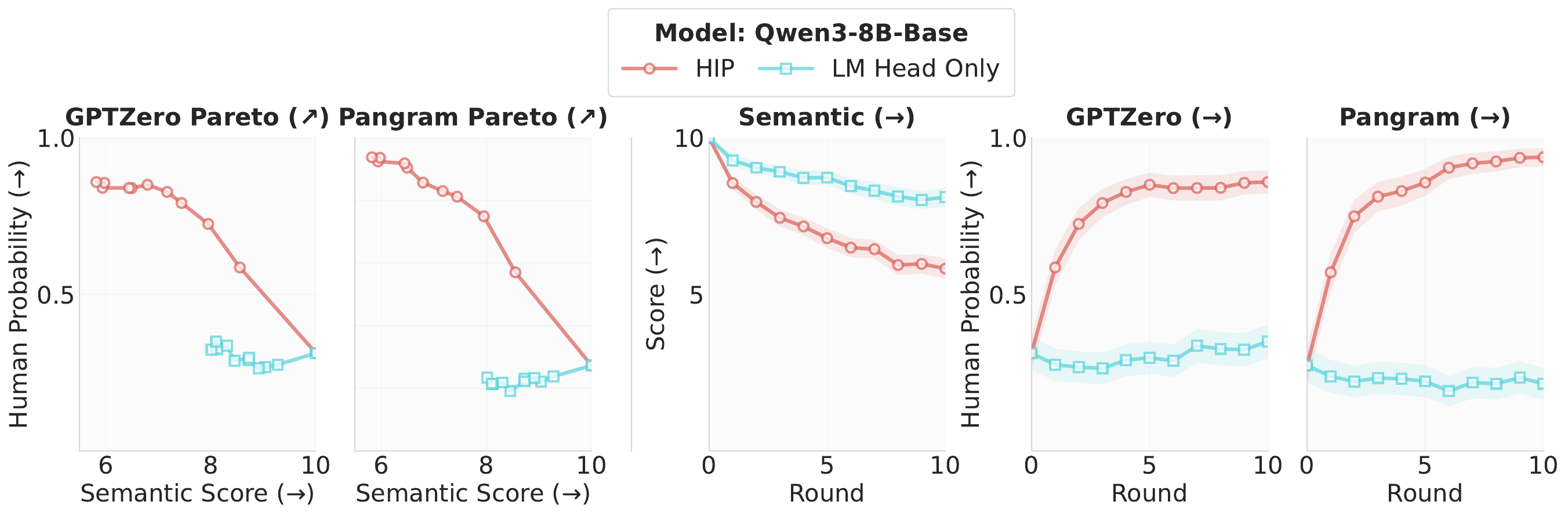}
    \caption{Standard HIP versus output-layer-only adaptation. Top: Llama3-8B-Base. Bottom: Qwen3-8B-Base. In each panel, the first two subplots show the GPTZero and Pangram Pareto frontiers for the trade-off between semantic preservation and human-likeness, while the last three show semantic score, GPTZero human probability, and Pangram human probability over the rounds, respectively. Shaded areas indicate $95\%$ confidence intervals. Output-layer-only adaptation underperforms standard HIP in both families, and especially on Qwen3-8B it fails to produce a useful semantic-evasion trade-off.}
    \label{fig:app_lm_head}
\end{figure}

As \cref{fig:app_lm_head} shows, output-layer-only adaptation does not reproduce the main HIP trade-off. For Llama3-8B, the detector curves move in the right direction, but only at the cost of a much sharper drop in semantic preservation than under standard HIP. For Qwen3-8B, the failure is stronger: semantic scores remain relatively high compared to that in Llama3-8B, but detector-assigned human probability stays far below the HIP curve on both GPTZero and Pangram.

Taken together, these results suggest that the main HIP effect cannot be explained as a simple output-layer transformation, such as reweighting or suppressing a small set of suspicious tokens. At least for the models studied here, successful humanization appears to require changes deeper in the representation pathway than the final logit layer alone.

\newpage
\section{Qualitative Examples}
\label{app:qual_examples}
To help interpret the quantitative metrics in the main paper, this appendix presents matched qualitative examples for HIP on Llama3-8B and for all four baseline methods: Simple Paraphrase, DIPPER, SilverSpeak, and StealthRL. The five examples appear in the same source-text order for every method and are sampled in an unbiased manner. For HIP, Simple Paraphrase, DIPPER, and StealthRL, each example shows the original AI text together with the first, second, and tenth rewrites, along with their semantic, GPTZero, and Pangram scores. SilverSpeak is a single-pass method, so its examples use a one-row layout with the original text and the transformed output. For readability in the PDF, the displayed SilverSpeak outputs are ASCII-normalized renderings of the homoglyph-substituted text, while the reported scores are computed on the original attacked text. 

\input{appendix_qual_examples.tex}

\end{document}

%% file: figures/method/hip_method.tex
% HIP method figure.
% Color semantics: blue = human-like, orange = AI-like.
% Requires in the preamble of the including document:
%   \usepackage{tikz}
%   \usetikzlibrary{arrows.meta,positioning,calc,shapes.geometric,decorations.pathmorphing,decorations.pathreplacing}
\begin{tikzpicture}[
    font=\small,
    /utils/exec={%
        \colorlet{humancol}{blue!55!black}%
        \colorlet{aicol}{orange!75!black}%
    },
    box/.style={
        draw, rounded corners=2.5pt, thick,
        minimum width=19mm, minimum height=12mm,
        align=center
    },
    model/.style={box, fill=blue!8, draw=humancol},
    textbox/.style={box, fill=orange!12, draw=aicol},
    humanbox/.style={box, fill=blue!8, draw=humancol},
    arr/.style={-{Latex[length=2.4mm,width=2.0mm]}, thick, draw=humancol},
    aiarr/.style={-{Latex[length=2.4mm,width=2.0mm]}, thick, draw=aicol},
    loop/.style={-{Latex[length=2.4mm,width=2.0mm]}, thick, draw=humancol!50!aicol},
    onlbl/.style={font=\footnotesize\itshape, align=center, inner sep=1.5pt},
    stagelbl/.style={font=\footnotesize\bfseries, align=center}
]
    % Layout coordinates
    \def\xbase{0.0}
    \def\xpara{5.0}
    \def\xtext{10.0}
    \def\boxhalf{0.95}
    \def\ytop{2.1}
    \def\xhuman{2.5}
    \def\xai{7.5}
    \pgfmathsetmacro{\xstagetwo}{((\xbase - \boxhalf) + \xpara) / 2}
    \pgfmathsetmacro{\xstagethree}{(\xpara + (\xtext + \boxhalf)) / 2}

    % Horizontal dashed divider separating Stage 1 (top row) from Stages 2/3.
    \draw[dashed, gray!75, thick] (-1.2, 1.4) -- (11.2, 1.4);
    % Vertical dashed divider through the Paraphraser (Stages 2/3 only).
    \draw[dashed, gray!75, thick] (\xpara, -1.6) -- (\xpara, 1.4);

    % --- Stage 1 row: Data collection ---
    \node[humanbox, minimum height=8mm] (htext) at (\xhuman, \ytop) {Human text $\textcolor{humancol}{h_i}$};
    \node[textbox,  minimum height=8mm] (atext) at (\xai,    \ytop) {AI text $\textcolor{aicol}{a_i}$};
    \draw[aiarr] (htext) -- node[onlbl, above=1pt, text=aicol] {AI paraphrase}
                            (atext);
    \node[stagelbl] at (\xpara, 3.2) {Stage 1\\Data collection};

    % --- Stages 2 and 3 row ---
    \node[model]   (base) at (\xbase, -0.6) {Base\\Model};
    \node[model]   (para) at (\xpara, -0.6) {Paraphraser\\$\mathcal{M}_{\mathrm{para}}$};
    \node[textbox] (text) at (\xtext, -0.6) {Text\\$x^{(0)}\dots x^{(N)}$};

    % Stage 2 / 3 headers
    \node[stagelbl] at (\xstagetwo,  0.8)
        {Stage 2\\Minimal fine-tuning};
    \node[stagelbl] at (\xstagethree, 0.8)
        {Stage 3\\Iterative paraphrasing};

    % Stage 2 arrow: base -> paraphraser
    \draw[arr] (base) -- node[onlbl, above=1pt] {$\textcolor{aicol}{a_i} \to \textcolor{humancol}{h_i}$}
                          node[onlbl, below=1pt] {preserving \textcolor{humancol}{humanness}}
                          (para);

    % Stage 3 loop arrows. Counterclockwise: feedback across the top, generation
    % across the bottom. Both share the mid-color (blue mixed with orange).
    \draw[loop, draw=aicol!67!humancol] ([yshift=1mm]text.west) --
        node[onlbl, above=1pt, text=aicol!67!humancol] {paraphrase $x^{(t)}$}
        ([yshift=1mm]para.east);
    \draw[loop, draw=aicol!33!humancol] ([yshift=-1mm]para.east) --
        node[onlbl, below=1pt, text=aicol!33!humancol] {for $x^{(t+1)}$}
        ([yshift=-1mm]text.west);
\end{tikzpicture}

%% file: appendix_qual_examples.tex
\subsection{HIP on Llama3-8B}
\noindent\textbf{Example 1.}\par\vspace{0.18em}
{\renewcommand{\qualbodyfont}{\scriptsize}%
\setlength{\qualpanelheight}{2.20in}
\noindent\qualexamplepanel{Original AI Text}{MAGE / XSum}{$10$ (ref.)}{$99.9$\%}{$0.0$\%}{Amputee football is played all over the world but it is in its infancy in Scotland. Amputee football is a form of the game that is adapted for people who have had one or both legs amputated. The Scottish Amputee Football Association (SAFA) is the governing body for the sport in Scotland and is working to set up a new league. There are currently only two teams in Scotland - Glasgow Rangers and Heart of Midlothian - but SAFA is hopeful that more clubs will join in the future. Amputee football is a fast-paced and physically demanding sport, but it is also great fun and can be enjoyed by people of all ages and abilities. If you are interested in finding out more about amputee football in Scotland, or if you would like to join a team, please contact SAFA.}\hfill
\qualexamplepanel{HIP (Llama3-8B) Round 1}{MAGE / XSum}{$10$}{$99.9$\%}{$0.0$\%}{Amputee football is a sport that is played all over the world, but it is still in its infancy in Scotland. Amputee football is a variation of the sport that has been adapted for people who have had their legs amputated. The Scottish Amputee Football Association (SAFA) is the governing body for amputee football in Scotland and is in the process of setting up a new league. There are currently only two teams in Scotland (Glasgow Rangers and Heart of Midlothian) but SAFA hopes that more teams will join in the future. Amputee football is a fast paced, physically demanding sport, but it is also a lot of fun and can be enjoyed by people of all ages and abilities. If you are interested in finding out more about amputee football in Scotland or if you are interested in joining a team, please contact SAFA.}

\vspace{0.45em}

\setlength{\qualpanelheight}{2.20in}
\qualexamplepanel{HIP (Llama3-8B) Round 2}{MAGE / XSum}{$10$}{$99.9$\%}{$100.0$\%}{Amputee football is a sport that is played all over the world, but it is in its infancy in Scotland. Amputee football is a variation of the game that has been adapted for people who have had their legs amputated. The Scottish Amputee Football Association (SAFA) is the governing body for amputee football in Scotland and are currently setting up a new league. There are currently only 2 teams in Scotland, Glasgow Rangers and Heart of Midlothian, but SAFA are hoping to get more teams in the future. Amputee football is a fast and physically demanding sport, but it is also a lot of fun and can be played by people of all ages and abilities. If you are interested in finding out more about amputee football in Scotland or if you are interested in joining a team please get in touch with SAFA.}\hfill
\qualexamplepanel{HIP (Llama3-8B) Round 10}{MAGE / XSum}{$9$}{$100.0$\%}{$100.0$\%}{Amputee Football is played all over the world, including Scotland.
Amputee football is just football, but it's amputees playing the game. At the moment, the Scottish Amputee Football Association (SAFA) are running a league for this game. There are only 2 teams in Scotland at the moment, Glasgow Rangers and Heart of Midlothian. The Scottish Amputee Football Association are trying to run more teams for this game as soon as possible. Amputee football is a fast, physical game. But it's a fun game and can be played by anyone of any age or ability. If you would like any more information on Amputee Football in Scotland, or if you would like to sign up to one of our teams, then please contact the Scottish Amputee Football Association.}
}
\clearpage

\noindent\textbf{Example 2.}\par\vspace{0.18em}
{\renewcommand{\qualbodyfont}{\scriptsize}%
\setlength{\qualpanelheight}{3.40in}
\noindent\qualexamplepanel{Original AI Text}{RAID / Abstracts}{$10$ (ref.)}{$0.0$\%}{$0.0$\%}{In this paper, we investigate the eigenvalues of non-Hermitian random matrices and the Brown measure of non-normal operators using a Hermitian reduction and linearization method. We show that the eigenvalues of non-Hermitian random matrices can be reduced to those of their Hermitian counterparts, and that the Brown measure of non-normal operators can be expressed in terms of the eigenvalue distribution of a associated linear operator. Our approach relies on a novel application of the perturbation theory of eigenvalues and the moment method, and provides a new perspective on the relationship between non-Hermitian and Hermitian systems.

We demonstrate the power of our method by computing the eigenvalue distribution of several classes of non-Hermitian random matrices, including the Gaussian Unitary Ensemble (GUE), the Gaussian Orthogonal Ensemble (GOE) and the Wishart matrix ensemble. Furthermore, we apply our results to study the Brown measure of non-normal operators, which has applications in various fields such as signal processing, control theory, and random matrix theory.

Our work provides a significant generalization of previous results in the field, and sheds light on the universal properties of non-Hermitian systems. The techniques developed here are expected to have far-reaching implications for the study of non-Hermitian phenomena in physics, engineering and other areas where complex systems are prevalent.}\hfill
\qualexamplepanel{HIP (Llama3-8B) Round 1}{RAID / Abstracts}{$10$}{$0.0$\%}{$0.0$\%}{In this paper, we study the eigenvalues of non-Hermitian random matrices and the Brown measure of non-normal operators by using a technique of Hermitian reduction and linearization. We show that the eigenvalues of non-Hermitian random matrices can be reduced to those of their Hermitian counterparts, and the Brown measure of non-normal operators can be expressed by the eigenvalue distribution of a linear operator. Our method is based on a new application of the perturbation theory of eigenvalues and the moment method, and provides a new angle to the relation between non-Hermitian and Hermitian systems. We apply our method to calculate the eigenvalue distribution of some non-Hermitian random matrices, including the Gaussian Unitary Ensemble (GUE), the Gaussian Orthogonal Ensemble (GOE), and the Wishart matrix ensemble. Furthermore, we apply our results to study the Brown measure of non-normal operators, which has applications in various areas such as signal processing, control theory, and random matrix theory. Our results provide a significant generalization of the previous works and provide a deeper understanding of the universal properties of non-Hermitian systems. Our approach has significant implications for understanding non-Hermitian phenomena in physics, engineering, and other areas where complex systems occur.}

\vspace{0.45em}

\setlength{\qualpanelheight}{3.1in}
\qualexamplepanel{HIP (Llama3-8B) Round 2}{RAID / Abstracts}{$9$}{$0.0$\%}{$0.0$\%}{In this paper, we study the eigenvalues of non-Hermitian random matrices and the Brown measure of non-normal operators by a method of Hermitian reduction and linearization. We show that the eigenvalues of non-Hermitian random matrices can be reduced to the eigenvalues of their Hermitian counterparts and the Brown measure of non-normal operators can be expressed in terms of the eigenvalue distribution of a linear operator. Our method is based on a new application of the perturbation theory of eigenvalues and the moment method, and shows a new direction to the relationship between non-Hermitian and Hermitian systems. Our method is applied to evaluate the eigenvalue distribution of some non-Hermitian random matrices such as Gaussian Unitary Ensemble (GUE), Gaussian Orthogonal Ensemble (GOE) and Wishart matrix ensemble. In addition, our results are applied to evaluate the Brown measure of non-normal operators, which has many applications in fields including signal processing, control theory and random matrix theory. Our results give a substantial extension of the known results, and provide a new insight into the universal properties of non-Hermitian systems. Our results have significant implications on the studies of non-Hermitian phenomenon in physics, engineering and other fields where complex systems exist.}\hfill
\qualexamplepanel{HIP (Llama3-8B) Round 10}{RAID / Abstracts}{$8$}{$99.9$\%}{$0.0$\%}{The eigenvalues of non-Hermitian random matrices and the Brown measure of a non-normal operator are studied via the reduction to Hermitian matrices and linearization methods. The results are applicable to the study of the eigenvalues of non-Hermitian random matrices via reduction to Hermitian matrices and the study of the Brown measure of a non-normal operator via linearization of the eigenvalues of a linear operator. In addition, this method is applicable to the study of the perturbation of the eigenvalues and moment theory of non-Hermitian and Hermitian systems, which is our main application. In particular, this method is used in studying the eigenvalues of non-Hermitian random matrices, such as Gaussian Unitary Ensemble (GUE), Gaussian Orthogonal Ensemble (GOE) and Wishart matrix. Besides, the distribution of the Brown measure of a non-normal operator is studied and it has important applications in signal processing, control theory and random matrix theory. This result is much improved and can provide new ideas and insights to the study of universal properties of non-Hermitian systems. These results are used in the non-Hermitian in physics, engineering, complex systems and other fields.}
}
\clearpage

\noindent\textbf{Example 3.}\par\vspace{0.18em}
{\renewcommand{\qualbodyfont}{\scriptsize}%
\setlength{\qualpanelheight}{2.4in}
\noindent\qualexamplepanel{Original AI Text}{RAID / Books}{$10$ (ref.)}{$0.0$\%}{$0.0$\%}{In "O Jerusalem", a young woman named Rachel returns to her hometown of Jerusalem after years of living abroad. She is drawn back by the promise of rebuilding her family's home, which was destroyed during the Israeli-Palestinian conflict. However, upon her arrival, Rachel finds that the city is more divided than ever, and tensions are running high between Jews and Arabs. As she navigates the complex social and political landscape, Rachel becomes involved with a group of activists who are fighting for peace and reconciliation between the two communities. Along the way, she encounters old friends, new loves, and bitter enemies, all while trying to find her own place in a city that is both sacred and profane. As the conflict escalates and the stakes grow higher, Rachel must decide where her loyalties truly lie and what she is willing to sacrifice for the sake of peace.}\hfill
\qualexamplepanel{HIP (Llama3-8B) Round 1}{RAID / Books}{$9$}{$0.0$\%}{$0.0$\%}{'O Jerusalem' is the story of a young woman, Rachel, who returns to her hometown of Jerusalem after many years living abroad. She is drawn back by the prospect of rebuilding her family's home, which was destroyed during the Israeli-Palestinian conflict. But on her arrival, she finds that Jerusalem has never been more divided, and tensions are high between the Jews and Arabs. As she becomes involved in the political and social struggles of the city, Rachel finds herself drawn to a group of activists working for peace and reconciliation between the two communities. Along the way, she makes new friends, falls in love, and encounters bitter enemies, as she tries to find her own place in this city of both sacred and profane beauty. As the conflict escalates, and the stakes get higher, Rachel must decide where her true loyalties lie, and what she is willing to sacrifice for the sake of peace.}

\vspace{0.45em}

\setlength{\qualpanelheight}{2.4in}
\qualexamplepanel{HIP (Llama3-8B) Round 2}{RAID / Books}{$9$}{$0.0$\%}{$0.0$\%}{The book tells the story of a young girl named Rachel who, after spending many years living abroad, returns to her native Jerusalem to rebuild her family's home which was destroyed during the Israeli-Palestinian conflict. When Rachel returns to Jerusalem, she finds that the city is at its most divided and the tension between Jews and Arabs is higher than ever. As she becomes involved with the political and social issues in the city, she is drawn to a group of activists who are trying to bring peace and reconciliation between the two groups. Along the way, Rachel makes new friends, finds love, and enemies, as she tries to find her place in the city, beautiful in its sacred and profane aspects. The conflict increases and the stakes are raised, and Rachel must decide which side she is on and what she is willing to sacrifice for the peace.}\hfill
\qualexamplepanel{HIP (Llama3-8B) Round 10}{RAID / Books}{$6$}{$100.0$\%}{$100.0$\%}{Rachel is a little girl that has just come back to live after spending years of her life in another country. She came back to rebuild her family's house that was destroyed by the Israeli-Palestinian War. When Rachel gets to her "holy and profane city" of Jerusalem, she is immediately told her city has never been more divided and the tensions between the Jewish and Arab residents is at an all time high. Rachel finds out everything that is happening in the city and is involved in every political and social event that is taking place in the city. Rachel learns about some people that wanted to create peace between the two countries. Rachel finds new friends, falls in love, makes enemies. She also learns how to accept that she lives in her "holy and profane city." It's not easy. The tensions between the two countries has never been higher and Rachel has to learn who she is and what she is willing to give up in order for there to be peace.}
}
\clearpage

\noindent\textbf{Example 4.}\par\vspace{0.18em}
{\renewcommand{\qualbodyfont}{\scriptsize}%
\setlength{\qualpanelheight}{3.7in}
\noindent\qualexamplepanel{Original AI Text}{MAGE / TLDR}{$10$ (ref.)}{$0.0$\%}{$0.0$\%}{Google has decided to discontinue Google Cloud Print after the year 2020, joining a list of other products which have been put to rest in recent years. This includes Google Reader and Google Inbox. Google Cloud Print, which was launched in 2010, is a service that allows users to print documents from anywhere, as long as they are connected to the internet. This means that a document can be printed from a smartphone, tablet or laptop without any need for cables or drivers. Google has made it clear that the service will continue to work until the end of 2020, but will no longer receive any software updates. Furthermore, from January 1, 2021, it will stop working entirely. The decision to discontinue Google Cloud Print comes as the company looks to focus on more profitable ventures. In a statement, Google said that it was discontinuing the service "to improve our overall cloud printing experience." The announcement has come as a surprise to many users, who see the service as an essential tool for printing from mobile devices. However, Google has suggested that there are a number of other cloud printing services available that will be able to fill the void left by Google Cloud Print. For those who are still using Google Cloud Print, Google has suggested that they begin transferring to another service before the year 2020 comes to a close. This will ensure that they are not left without an essential printing service. Overall, the discontinuation of Google Cloud Print is a reflection of the changing landscape of the technology industry. As companies look to streamline their operations and focus on more profitable ventures, it is inevitable that some beloved products and services will be phased out.}\hfill
\qualexamplepanel{HIP (Llama3-8B) Round 1}{MAGE / TLDR}{$9$}{$99.9$\%}{$0.0$\%}{Google has announced that it will be discontinuing Google Cloud Print after 2020. Other recent products that have been discontinued include Google Reader and Google Inbox. Google Cloud Print was a service introduced in 2010 that allowed users to print documents from anywhere with an internet connection. It allowed users to print from smartphones, tablets, and laptops without the need for a cable or driver. Google has announced that the service will still be available until 2020, but will no longer be updated. Starting on January 1, 2021, the service will no longer function. Google is discontinuing Google Cloud Print in order to focus on more profitable ventures. "After carefully reviewing the Google Cloud Print landscape and usage, we're discontinuing the service in order to focus on projects that bring greater value to our users," wrote Google in a statement. Google Cloud Print has been seen as an essential service for printing documents from mobile devices. The announcement has been surprising to many users, but Google has provided a list of other cloud printing services that will be able to replace it. Users who are still using Google Cloud Print will need to start transitioning to another service before 2020 in order to avoid any issues. Overall, this decision shows how the technology industry is changing. As companies look to streamline and focus on more profitable activities, it is inevitable that products and services that are beloved will have to be discontinued.}

\vspace{0.45em}

\setlength{\qualpanelheight}{3.7in}
\qualexamplepanel{HIP (Llama3-8B) Round 2}{MAGE / TLDR}{$9$}{$100.0$\%}{$0.0$\%}{Google has announced that it will be discontinuing Google Cloud Print. Other products that were recently discontinued by Google include Google Reader and Google Inbox. Google Cloud Print was a service introduced in 2010 that allowed users to print their documents from anywhere in the world with an Internet connection. It let users print documents directly from their smartphone, tablet, or laptop without the need for a cable or driver. Google has announced that the service will still be available until 2020 but will no longer be updated. Starting on January 1, 2021, the service will no longer function. Google has decided to discontinue the Google Cloud Print service due to its focus on more profitable endeavors. "After carefully reviewing the Google Cloud Print landscape and usage, we're discontinuing the service in order to focus on projects that bring greater value to our users," said Google in a statement. Google Cloud Print was a valuable service for those who wanted to print their documents from their mobile devices. Many users have been shocked by the announcement, but Google has provided a list of other cloud printing services that will be able to replace Google Cloud Print. Users will have to start transitioning to the other service before 2020 if they still use Google Cloud Print in order to avoid any problems. The technology industry is changing, and this is something that has been seen for some time. As the industry is trying to streamline its operations and focus more on more profitable ventures, it is only a matter of time before beloved products and services are discontinued.}\hfill
\qualexamplepanel{HIP (Llama3-8B) Round 10}{MAGE / TLDR}{$7$}{$100.0$\%}{$100.0$\%}{Google Cloud Print will be shut down. Google has shut down two of its services recently. The services that Google has shut down are Google Reader and Google Inbox. Google Cloud Print launched in 2010. Google Cloud Print was a service that lets you print files from anywhere in the world as long as you have internet. Google Cloud Print allowed people to print documents to any printer without having to use wires or drivers. Google Cloud Print will still be running until 2020, but Google will no longer be developing it. Google Cloud Print will no longer be working after January 1, 2021. Google shut down Google Cloud Print because it was making more money with other products. After "careful consideration of the Google Cloud Print ecosystem and usage, we have decided to discontinue Google Cloud Print in order to focus on other projects that provide greater value to our users," Google said in a statement. Google Cloud Print was a great service for people who needed to print from their phone. Many people were shocked that Google was shutting down Google Cloud Print. Google has a list of other cloud printing services. People will have to start using the new services in 2020 if they still use Google Cloud Print. Technology is always like this. Businesses are trying to be more efficient and are focusing on the things that are more profitable for them. Eventually, we will just start shutting down the things we like.}
}
\clearpage

\noindent\textbf{Example 5.}\par\vspace{0.18em}
{\renewcommand{\qualbodyfont}{\scriptsize}%
\setlength{\qualpanelheight}{3.4in}
\noindent\qualexamplepanel{Original AI Text}{MAGE / SQuAD}{$10$ (ref.)}{$100.0$\%}{$100.0$\%}{The Panthers offense, which led the NFL in scoring (500 points), was loaded with talent, boasting six Pro Bowl selections. Pro Bowl quarterback Cam Newton had one of his best seasons, completing 67.9 percent of his passes (296 of 432) for 3,837 yards and 35 touchdowns, with just 10 interceptions. Newton was the first quarterback in NFL history to throw for more than 30 touchdowns and rush for 10 in a single season. He had 10 rushing touchdowns in 2015, the most by a quarterback since he had 14 in 2011. Newton's 35 passing touchdowns set a Panthers franchise record. His favorite target was tight end Greg Olsen, who led the team in receptions (77) and receiving yards (1,104) and had seven touchdown catches. It was the third consecutive season that Olsen led the Panthers in receiving yards. He is one of six tight ends in NFL history with three consecutive 1,000-yard receiving seasons. While the offense was explosive, the defense was stingy. The Panthers finished the season ranked second in the NFL in both total defense and points allowed per game (19.3). Defensive end Jared Allen led the team with 10.5 sacks, and linebacker Thomas Davis had a career-high 115 tackles, two forced fumbles and a career-best four interceptions, returning one for a touchdown. Cornerback Josh Norman finished the season with four interceptions and 18 passes defensed.}\hfill
\qualexamplepanel{HIP (Llama3-8B) Round 1}{MAGE / SQuAD}{$7$}{$100.0$\%}{$100.0$\%}{The Panthers offense, led by a first-year coach, was the top-scoring unit in the NFL, finishing the year with 500 points scored. They had six Pro Bowl selections. Pro Bowl quarterback Cam Newton had his best season to date, completing 67.9 (296297432) of his passes for 3,837 yards, 35 touchdowns, and 10 interceptions. He was the first player in NFL history to throw for more than 30 touchdowns and run for 10 touchdowns in the same season. He had 10 rushing touchdowns in 2015, the most by a quarterback since his 14 rushing touchdowns in 2011. His 35 passing touchdowns set a Panthers franchise record. His top target was tight end Greg Olsen, who led the team in receptions (77) and receiving yards (1,104) and had seven touchdowns. Olsen was the team leader in receiving yards for the third consecutive season. He is one of only six NFL tight ends to ever have three consecutive 1,000-yard seasons. While the offense was explosive, the defense was stingy. The Panthers finished the season second in the NFL in both total defense and points allowed per game (19.3). Defensive end Jared Allen led the team with 10.5 sacks, and linebacker Thomas Davis had a career-high 115 tackles, two forced fumbles, and a career-best four interceptions, returning one of the interceptions for a touchdown. Cornerback Josh Norman finished the season with four interceptions and 18 passes defended.}

\vspace{0.45em}

\setlength{\qualpanelheight}{3.4in}
\qualexamplepanel{HIP (Llama3-8B) Round 2}{MAGE / SQuAD}{$8$}{$99.9$\%}{$100.0$\%}{The team's offense, led by a first-year head coach, was the most potent scoring unit in the NFL in 2015. The team scored 500 total points and was represented with six players at the Pro Bowl. Panthers quarterback Cam Newton had a career year with a completion percentage of 67.9 percent on 296 passes (296297432), 3,837 passing yards, 35 touchdown passes, and 10 interceptions. He was the first player in NFL history to throw 30 touchdown passes and rush for 10 touchdowns in the same season. Newton had 10 rushing touchdowns in 2015, the most by a quarterback since Newton's 14 rushing touchdowns in 2011. His 35 touchdown passes set a new Panthers franchise record. Newton's favorite target was tight end Greg Olsen, who led the team in both receptions (77) and receiving yards (1,104), and touchdowns (7). Olsen has led the team in receiving yards each of the last three years. He is only one of six players in NFL history to have three consecutive 1,000-yard seasons. While the Panthers' offense was explosive, the defense was opportunistic. The Panthers ended the season ranked second in the NFL in both total defense and points allowed (19.3 points per game). Defensive end Jared Allen led the team with 10.5 sacks and linebacker Thomas Davis had a career-high 115 tackles, 2 forced fumbles, and a career-high 4 interceptions. Davis also had a career-best one interception returned for a touchdown. Cornerback Josh Norman finished the season with four interceptions and 18 passes defended.}\hfill
\qualexamplepanel{HIP (Llama3-8B) Round 10}{MAGE / SQuAD}{$3$}{$100.0$\%}{$100.0$\%}{2015 Carolina Panthers Offensive Stats
500 Points - 1st in NFL In Cam Newton's 1st Season as Head Coach. 
 6 players went to 2016 Pro Bowl. Cam Newton had one of the best seasons of his Career. Cam Newton completed 67.9 of passes out of 29732 pass attempts, had 3837 Pass Yards, and threw for 35 touchdowns, threw 10 interceptions and 10 passes dropped by wide receivers. Cam Newton was the 1st player in NFL History to throw 30 passing touchdowns and rush for 10 rushing touchdowns. Cam Newton also had 10 rushing touchdowns the most by a quarterback since 2011 (Newton had 14 Rushing TD's in 2011) Cam Newton had the most touchdown passes in a season of any Panther ever. His primary target was tight end Greg Olsen. Greg Olsen was the Panthers top receiver for 3 straight seasons. Greg Olsen had the most receptions (77) yards (1104) and touchdowns (7) in his career. Greg Olsen was one of only 6 players in NFL History to catch 3 straight 1000 yard receiving seasons. The Carolina Panthers were very good on both sides of the ball. 2nd in the NFL in fewest points allowed (19.3), and fewest yards allowed in 2015. Defensive End Jared Allen had 10.5 sacks in 2015. Linebacker Thomas Davis had 115 tackles the most on the team, 2nd in the NFL, 2 forced fumbles and 4 interceptions (career high) Thomas Davis 1 of 4 interceptions was returned for a touchdown, Career high. Cornerback Josh Norman led the team in interceptions with 4 and Passes defensed with 18}
}
\clearpage

\subsection{Simple Paraphrase}
\noindent\textbf{Example 1.}\par\vspace{0.18em}
{\renewcommand{\qualbodyfont}{\scriptsize}%
\setlength{\qualpanelheight}{2.20in}
\noindent\qualexamplepanel{Original AI Text}{MAGE / XSum}{$10$ (ref.)}{$99.9$\%}{$0.0$\%}{Amputee football is played all over the world but it is in its infancy in Scotland. Amputee football is a form of the game that is adapted for people who have had one or both legs amputated. The Scottish Amputee Football Association (SAFA) is the governing body for the sport in Scotland and is working to set up a new league. There are currently only two teams in Scotland - Glasgow Rangers and Heart of Midlothian - but SAFA is hopeful that more clubs will join in the future. Amputee football is a fast-paced and physically demanding sport, but it is also great fun and can be enjoyed by people of all ages and abilities. If you are interested in finding out more about amputee football in Scotland, or if you would like to join a team, please contact SAFA.}\hfill
\qualexamplepanel{Simple Paraphrase Round 1}{MAGE / XSum}{$9$}{$0.0$\%}{$0.0$\%}{Across the globe, amputee football has gained popularity, but it is still in its early stages of development in Scotland. This modified version of the game caters to individuals who have undergone one or both leg amputations. The Scottish Amputee Football Association (SAFA) is actively working to establish a new league, currently with only two teams, Glasgow Rangers and Heart of Midlothian, but anticipating more clubs to join soon. Despite the physical demands of the game, amputee football is an exhilarating and inclusive sport suitable for people of all ages and abilities. Those interested in learning more about amputee football in Scotland or wanting to join a team can reach out to SAFA.}

\vspace{0.45em}

\setlength{\qualpanelheight}{2.20in}
\qualexamplepanel{Simple Paraphrase Round 2}{MAGE / XSum}{$9$}{$99.1$\%}{$0.0$\%}{Around the world, amputee football is gaining traction, but Scotland is still in the process of building its scene. This adapted version of the game is specifically designed for individuals who have undergone one or both leg amputations. The Scottish Amputee Football Association is working towards establishing a new league, with two founding teams - Glasgow Rangers and Heart of Midlothian - and expects more clubs to join soon. Despite the physical demands of the sport, amputee football offers an exciting and inclusive experience for people of all ages and abilities. Those curious about amputee football in Scotland or interested in joining a team can contact the Scottish Amputee Football Association for more information.}\hfill
\qualexamplepanel{Simple Paraphrase Round 10}{MAGE / XSum}{$8$}{$0.0$\%}{$0.0$\%}{As recognition of adaptive football grows globally, Scotland is establishing its own adaptive football network, providing a platform for individuals with leg amputations to play the game. This modified version of football breaks down physical barriers, allowing players with leg amputations to fully participate and be an integral part of the action. The Scottish Amputee Football Association is driving an initiative to launch a new league, initially featuring well-known clubs like Glasgow Rangers and Heart of Midlothian, with more teams expected to join soon. Despite physical limitations, adaptive football offers an exhilarating and inclusive experience, bridging the gap among individuals of all ages and abilities, and cultivating strong bonds and a sense of belonging among players.}
}
\clearpage

\noindent\textbf{Example 2.}\par\vspace{0.18em}
{\renewcommand{\qualbodyfont}{\scriptsize}%
\setlength{\qualpanelheight}{3.40in}
\noindent\qualexamplepanel{Original AI Text}{RAID / Abstracts}{$10$ (ref.)}{$0.0$\%}{$0.0$\%}{In this paper, we investigate the eigenvalues of non-Hermitian random matrices and the Brown measure of non-normal operators using a Hermitian reduction and linearization method. We show that the eigenvalues of non-Hermitian random matrices can be reduced to those of their Hermitian counterparts, and that the Brown measure of non-normal operators can be expressed in terms of the eigenvalue distribution of a associated linear operator. Our approach relies on a novel application of the perturbation theory of eigenvalues and the moment method, and provides a new perspective on the relationship between non-Hermitian and Hermitian systems.

We demonstrate the power of our method by computing the eigenvalue distribution of several classes of non-Hermitian random matrices, including the Gaussian Unitary Ensemble (GUE), the Gaussian Orthogonal Ensemble (GOE) and the Wishart matrix ensemble. Furthermore, we apply our results to study the Brown measure of non-normal operators, which has applications in various fields such as signal processing, control theory, and random matrix theory.

Our work provides a significant generalization of previous results in the field, and sheds light on the universal properties of non-Hermitian systems. The techniques developed here are expected to have far-reaching implications for the study of non-Hermitian phenomena in physics, engineering and other areas where complex systems are prevalent.}\hfill
\qualexamplepanel{Simple Paraphrase Round 1}{RAID / Abstracts}{$9$}{$0.0$\%}{$0.0$\%}{Here is the paraphrased passage:

In this research, we explore the eigenvalues of non-Hermitian random matrices and the Brown measure of non-normal operators using a Hermitian reduction and linearization approach. We reveal that the eigenvalues of non-Hermitian random matrices can be reduced to those of their Hermitian counterparts, and the Brown measure of non-normal operators can be expressed in terms of the eigenvalue distribution of a related linear operator. Our method combines perturbation theory and the moment method to provide a fresh perspective on the connection between non-Hermitian and Hermitian systems.

We showcase the effectiveness of our approach by calculating the eigenvalue distribution for various classes of non-Hermitian random matrices, including GUE, GOE, and the Wishart matrix ensemble. We also apply our findings to study the Brown measure of non-normal operators, which has practical applications in signal processing, control theory, and random matrix theory.

Our work extends previous research and uncovers universal properties of non-Hermitian systems, with the developed techniques likely to have significant implications for understanding complex systems in physics, engineering, and other fields where non-Hermitian phenomena are prevalent.}

\vspace{0.45em}

\setlength{\qualpanelheight}{1.9in}
\qualexamplepanel{Simple Paraphrase Round 2}{RAID / Abstracts}{$7$}{$0.0$\%}{$0.0$\%}{Here is the paraphrased passage:

This research delves into the eigenvalues of non-Hermitian random matrices and the Brown measure of non-normal operators, leveraging a Hermitian reduction and linearization technique. We discovered that the eigenvalues of non-Hermitian random matrices can be equivalently viewed as those of their Hermitian counterparts, and the Brown measure of non-normal operators can be expressed in terms of the eigenvalue distribution of a related linear operator. Our method combines perturbation theory and the moment method to offer a novel perspective on the connection between non-Hermitian and Hermitian systems.}\hfill
\qualexamplepanel{Simple Paraphrase Round 10}{RAID / Abstracts}{$4$}{$0.0$\%}{$0.0$\%}{This study employs an innovative approach to examine the eigenvalues of non-Hermitian random matrices and the behavior of non-normal operators by converting the non-Hermitian matrix into a Hermitian one, simplifying the analysis process. The findings reveal that the eigenvalues of non-Hermitian random matrices exhibit similar patterns to those of Hermitian matrices, and the properties of non-normal operators can be comprehended by analyzing the distribution of eigenvalues of a linked linear operator. By merging perturbation theory and the moment method, this methodology offers a fresh perspective on the connection between non-Hermitian and Hermitian systems.}
}
\clearpage

\noindent\textbf{Example 3.}\par\vspace{0.18em}
{\renewcommand{\qualbodyfont}{\scriptsize}%
\setlength{\qualpanelheight}{2.2in}
\noindent\qualexamplepanel{Original AI Text}{RAID / Books}{$10$ (ref.)}{$0.0$\%}{$0.0$\%}{In "O Jerusalem", a young woman named Rachel returns to her hometown of Jerusalem after years of living abroad. She is drawn back by the promise of rebuilding her family's home, which was destroyed during the Israeli-Palestinian conflict. However, upon her arrival, Rachel finds that the city is more divided than ever, and tensions are running high between Jews and Arabs. As she navigates the complex social and political landscape, Rachel becomes involved with a group of activists who are fighting for peace and reconciliation between the two communities. Along the way, she encounters old friends, new loves, and bitter enemies, all while trying to find her own place in a city that is both sacred and profane. As the conflict escalates and the stakes grow higher, Rachel must decide where her loyalties truly lie and what she is willing to sacrifice for the sake of peace.}\hfill
\qualexamplepanel{Simple Paraphrase Round 1}{RAID / Books}{$9$}{$0.0$\%}{$0.0$\%}{After years abroad, Rachel returns to Jerusalem, driven by the desire to restore her family's destroyed home, a symbol of a bygone era. However, her arrival coincides with a time of heightened tension between the Jewish and Arab communities, and she soon finds herself entangled in the daily struggles of a city torn apart by conflict. Rachel's path converges with a group of determined activists working towards peace, but her journey is also marked by encounters with old acquaintances, new relationships, and sworn enemies. As the situation becomes increasingly volatile, Rachel must weigh her loyalties and confront the personal costs of pursuing peace in a city that is both a sacred and a profane place.}

\vspace{0.45em}

\setlength{\qualpanelheight}{2.0in}
\qualexamplepanel{Simple Paraphrase Round 2}{RAID / Books}{$8$}{$0.0$\%}{$0.0$\%}{Rachel returns to Jerusalem after years away, drawn by a sense of duty to revive her family's ancestral home, a relic of a bygone era. However, her arrival coincides with a period of heightened anxiety between the Jewish and Arab communities, and she becomes entangled in the daily struggles of a city plagued by discord. As she navigates this treacherous landscape, Rachel's path intersects with a group of passionate advocates for peace, while also crossing paths with old friends, new acquaintances, and even sworn foes. As tensions escalate, Rachel must balance her allegiances and confront the personal costs of seeking harmony in a city that is both revered and profane.}\hfill
\qualexamplepanel{Simple Paraphrase Round 10}{RAID / Books}{$9$}{$98.4$\%}{$0.0$\%}{Rachel's ambitious endeavor to revive her family's Jerusalem property places her at the forefront of a sensitive and precarious situation, as she must tread carefully between the entrenched animosity and mistrust between the Jewish and Arab communities. As she works to build bridges, Rachel forges strong bonds with those striving for peaceful coexistence, rekindles old friendships, and even confronts former foes. Amidst escalating tensions, Rachel must delicately weigh her own allegiances against the personal sacrifices she must make to heal the deep-seated wounds of the past and bring the two communities closer together.}
}
\clearpage

\noindent\textbf{Example 4.}\par\vspace{0.18em}
{\renewcommand{\qualbodyfont}{\scriptsize}%
\setlength{\qualpanelheight}{3.7in}
\noindent\qualexamplepanel{Original AI Text}{MAGE / TLDR}{$10$ (ref.)}{$0.0$\%}{$0.0$\%}{Google has decided to discontinue Google Cloud Print after the year 2020, joining a list of other products which have been put to rest in recent years. This includes Google Reader and Google Inbox. Google Cloud Print, which was launched in 2010, is a service that allows users to print documents from anywhere, as long as they are connected to the internet. This means that a document can be printed from a smartphone, tablet or laptop without any need for cables or drivers. Google has made it clear that the service will continue to work until the end of 2020, but will no longer receive any software updates. Furthermore, from January 1, 2021, it will stop working entirely. The decision to discontinue Google Cloud Print comes as the company looks to focus on more profitable ventures. In a statement, Google said that it was discontinuing the service "to improve our overall cloud printing experience." The announcement has come as a surprise to many users, who see the service as an essential tool for printing from mobile devices. However, Google has suggested that there are a number of other cloud printing services available that will be able to fill the void left by Google Cloud Print. For those who are still using Google Cloud Print, Google has suggested that they begin transferring to another service before the year 2020 comes to a close. This will ensure that they are not left without an essential printing service. Overall, the discontinuation of Google Cloud Print is a reflection of the changing landscape of the technology industry. As companies look to streamline their operations and focus on more profitable ventures, it is inevitable that some beloved products and services will be phased out.}\hfill
\qualexamplepanel{Simple Paraphrase Round 1}{MAGE / TLDR}{$9$}{$0.0$\%}{$0.0$\%}{Google is retiring Google Cloud Print, a service that enabled users to print documents remotely, following in the footsteps of discontinued services like Google Reader and Google Inbox. Launched in 2010, Google Cloud Print allowed users to print from their devices without cables or drivers. Although the service will remain functional until the end of 2020, it will no longer receive updates, and starting January 2021, it will cease to function altogether. The decision to discontinue the service is part of Google's effort to concentrate on more profitable ventures. The company suggests that alternative cloud printing services can meet the demand, and advises users to switch before the year's end to avoid being left without a printing solution. This move reflects the ever-changing tech landscape, where companies optimize their operations by discontinuing non-profitable services to focus on more lucrative ones.}

\vspace{0.45em}

\setlength{\qualpanelheight}{2in}
\qualexamplepanel{Simple Paraphrase Round 2}{MAGE / TLDR}{$9$}{$0.0$\%}{$0.0$\%}{After a decade of service, Google is phasing out Google Cloud Print, a technology that let users print remotely without cables or drivers. Introduced in 2010, the service will remain active until December 2020 but will no longer receive updates or support starting January 2021. The decision is part of Google's strategy to prioritize profitable initiatives. Although alternative cloud printing solutions are available, users are advised to switch to them by the end of the year to avoid disruption. This move highlights the dynamic nature of the tech industry, where companies often discontinue underperforming services to focus on more lucrative ventures.}\hfill
\qualexamplepanel{Simple Paraphrase Round 10}{MAGE / TLDR}{$4$}{$15.3$\%}{$0.0$\%}{After a decade of pioneering driverless printing, Google has decided to discontinue its Cloud Print service. Although users will still be able to use the service until December 2020, Google will no longer provide updates or support from January 2021. This change will allow the company to shift its resources to more financially successful projects. To make the transition as seamless as possible, users are encouraged to switch to alternative cloud printing solutions by the end of the year, a common approach companies take when retiring underperforming services to focus on more profitable initiatives.}
}
\clearpage

\noindent\textbf{Example 5.}\par\vspace{0.18em}
{\renewcommand{\qualbodyfont}{\scriptsize}%
\setlength{\qualpanelheight}{3.4in}
\noindent\qualexamplepanel{Original AI Text}{MAGE / SQuAD}{$10$ (ref.)}{$100.0$\%}{$100.0$\%}{The Panthers offense, which led the NFL in scoring (500 points), was loaded with talent, boasting six Pro Bowl selections. Pro Bowl quarterback Cam Newton had one of his best seasons, completing 67.9 percent of his passes (296 of 432) for 3,837 yards and 35 touchdowns, with just 10 interceptions. Newton was the first quarterback in NFL history to throw for more than 30 touchdowns and rush for 10 in a single season. He had 10 rushing touchdowns in 2015, the most by a quarterback since he had 14 in 2011. Newton's 35 passing touchdowns set a Panthers franchise record. His favorite target was tight end Greg Olsen, who led the team in receptions (77) and receiving yards (1,104) and had seven touchdown catches. It was the third consecutive season that Olsen led the Panthers in receiving yards. He is one of six tight ends in NFL history with three consecutive 1,000-yard receiving seasons. While the offense was explosive, the defense was stingy. The Panthers finished the season ranked second in the NFL in both total defense and points allowed per game (19.3). Defensive end Jared Allen led the team with 10.5 sacks, and linebacker Thomas Davis had a career-high 115 tackles, two forced fumbles and a career-best four interceptions, returning one for a touchdown. Cornerback Josh Norman finished the season with four interceptions and 18 passes defensed.}\hfill
\qualexamplepanel{Simple Paraphrase Round 1}{MAGE / SQuAD}{$10$}{$7.6$\%}{$0.0$\%}{The Carolina Panthers' offense, which topped the NFL scoring charts with 500 points, was stacked with talent, boasting six Pro Bowl selections. Star quarterback Cam Newton enjoyed a standout season, finishing with a 67.9\% completion rate, 3,837 passing yards, and a franchise-record 35 touchdowns, along with just 10 interceptions. Newton achieved a historic feat, becoming the first quarterback to combine 30 touchdowns passing and 10 rushing in a single season. His 10 rushing touchdowns in 2015 marked the highest total by a quarterback since his own 14-touchdown performance in 2011. Meanwhile, tight end Greg Olsen emerged as Newton's go-to target, recording 77 receptions and 1,104 receiving yards, with seven touchdown catches. Olsen led the team in receiving yards for the third consecutive year, joining an elite group of tight ends with three consecutive 1,000-yard receiving seasons. While the offense was potent, the Panthers' defense shone just as brightly, ranking second in the NFL in both total defense and points allowed per game, with an impressive 19.3 average. Defensive leader Jared Allen topped the team with 10.5 sacks, while linebacker Thomas Davis boasted a career-best 115 tackles, two forced fumbles, and four interceptions, including a return for a touchdown. Cornerback Josh Norman concluded the season with four interceptions and 18 passes defensed.}

\vspace{0.45em}

\setlength{\qualpanelheight}{3.1in}
\qualexamplepanel{Simple Paraphrase Round 2}{MAGE / SQuAD}{$9$}{$0.0$\%}{$100.0$\%}{The Carolina Panthers' offense, which led the NFL in scoring with 500 points, was a force to be reckoned with, boasting an impressive six Pro Bowl selections. Star quarterback Cam Newton put up exceptional numbers, completing 67.9\% of his passes for 3,837 yards, and throwing a franchise-record 35 touchdowns, while only throwing 10 interceptions. Newton made history, becoming the first quarterback to combine 30 touchdowns passing and 10 rushing in a single season, and his 10 rushing touchdowns were the highest total by a quarterback since his own 14-touchdown performance in 2011. Tight end Greg Olsen emerged as Newton's favorite target, catching 77 passes for 1,104 yards and seven touchdowns, and became the first tight end in franchise history to record three consecutive 1,000-yard receiving seasons. The Panthers' defense, on the other hand, was nearly as impressive, ranking second in the league in both total defense and points allowed per game, with an impressive average of 19.3. Defensive leader Jared Allen led the team with 10.5 sacks, while linebacker Thomas Davis had a career-best 115 tackles, two forced fumbles, and four interceptions, including one that resulted in a touchdown. Cornerback Josh Norman finished the season with four interceptions and 18 passes defensed.}\hfill
\qualexamplepanel{Simple Paraphrase Round 10}{MAGE / SQuAD}{$8$}{$0.0$\%}{$0.0$\%}{The Carolina Panthers' offense made a significant impact this season, setting a league record with 500 points scored. The team's success was largely driven by the outstanding performance of six Pro Bowl players. Quarterback Cam Newton stood out, boasting an impressive 69\% completion rate, a franchise record 3,837 yards, and an impressive 35 touchdowns, all while minimizing interceptions to just 10. Newton achieved a remarkable milestone by becoming the first quarterback to reach 30 passing touchdowns and 10 rushing touchdowns in a single season. Wide receiver Greg Olsen proved to be Newton's go-to target, hauling in 77 passes for 1,104 yards and seven touchdowns, making him the first tight end in team history to achieve three consecutive 1,000-yard receiving seasons. The team's defense also played a crucial role, ranking second in the league in total defense and points allowed per game, with a relatively modest 19.3 points allowed per game. Notable defensive players like Jared Allen and Thomas Davis made significant contributions, with Allen recording 10.5 sacks and Davis boasting a career-high 115 tackles, two forced fumbles, and four interceptions, including one for a touchdown.}
}
\clearpage

\subsection{DIPPER}
\noindent\textbf{Example 1.}\par\vspace{0.18em}
{\renewcommand{\qualbodyfont}{\scriptsize}%
\setlength{\qualpanelheight}{2.20in}
\noindent\qualexamplepanel{Original AI Text}{MAGE / XSum}{$10$ (ref.)}{$99.9$\%}{$0.0$\%}{Amputee football is played all over the world but it is in its infancy in Scotland. Amputee football is a form of the game that is adapted for people who have had one or both legs amputated. The Scottish Amputee Football Association (SAFA) is the governing body for the sport in Scotland and is working to set up a new league. There are currently only two teams in Scotland - Glasgow Rangers and Heart of Midlothian - but SAFA is hopeful that more clubs will join in the future. Amputee football is a fast-paced and physically demanding sport, but it is also great fun and can be enjoyed by people of all ages and abilities. If you are interested in finding out more about amputee football in Scotland, or if you would like to join a team, please contact SAFA.}\hfill
\qualexamplepanel{DIPPER Round 1}{MAGE / XSum}{$3$}{$100.0$\%}{$100.0$\%}{So scout out this topic and go for a tour of a big city, there's some money to spend on the players' gear, but otherwise it's a relatively untapped market. Amputee football is played all over the world but in Scotland it's still in its infancy. The Scottish Amputee Football Association is the governing body in Scotland and it is working on establishing a league. There are only two teams in Scotland at present - Glasgow Rangers and Heart of Midlothian - but SAFA is hoping that other clubs will be keen to join. Amputee football is a fast-paced and demanding sport, but it's also a lot of fun and people of all ages and abilities are welcome to join. If you are interested in finding out more about amputee football in Scotland, or would like to join a team, then please contact SAFA.}

\vspace{0.45em}

\setlength{\qualpanelheight}{2.20in}
\qualexamplepanel{DIPPER Round 2}{MAGE / XSum}{$6$}{$100.0$\%}{$0.0$\%}{So scout out the topic and go on a city tour. There's money to be spent on the players' equipment, but otherwise it's a pretty raw deal. Amputee football is played all over the world, but in Scotland it is still in its infancy. The Scottish Amputee Association is the controlling body in Scotland, and they are trying to establish a league. There are only two teams in Scotland, Glasgow Rangers and Hearts of Midlothian, but they hope that more clubs will be interested in joining. Amputee football is a very intense and strenuous sport, but it is also very entertaining, and everyone of all ages and physical conditions is welcome. If you want to know more about amputee football in Scotland, or if you want to join a team, please contact the Scottish Amputee Association.}\hfill
\qualexamplepanel{DIPPER Round 10}{MAGE / XSum}{$6$}{$4.7$\%}{$100.0$\%}{And so we took up the project and began to work. We had money with which to buy the material; but it was difficult to get it together. The game we were promoting has been played in all countries since its invention; but in Scotland it is still in its infancy. The Scottish Amputee Association is the chief body promoting it; but it is working for a league of teams. At present only two of the clubs are represented--Rangers and Hearts--but it is hoped to get more. It is a rough, dangerous game, and at the same time it is amusing. It is suitable for all ages and all states of health. If you are interested in the progress of the Scottish Amputee Association, or if you wish to join a team, you can get in touch with it at the address below.}
}
\clearpage

\noindent\textbf{Example 2.}\par\vspace{0.18em}
{\renewcommand{\qualbodyfont}{\scriptsize}%
\setlength{\qualpanelheight}{3.40in}
\noindent\qualexamplepanel{Original AI Text}{RAID / Abstracts}{$10$ (ref.)}{$0.0$\%}{$0.0$\%}{In this paper, we investigate the eigenvalues of non-Hermitian random matrices and the Brown measure of non-normal operators using a Hermitian reduction and linearization method. We show that the eigenvalues of non-Hermitian random matrices can be reduced to those of their Hermitian counterparts, and that the Brown measure of non-normal operators can be expressed in terms of the eigenvalue distribution of a associated linear operator. Our approach relies on a novel application of the perturbation theory of eigenvalues and the moment method, and provides a new perspective on the relationship between non-Hermitian and Hermitian systems.

We demonstrate the power of our method by computing the eigenvalue distribution of several classes of non-Hermitian random matrices, including the Gaussian Unitary Ensemble (GUE), the Gaussian Orthogonal Ensemble (GOE) and the Wishart matrix ensemble. Furthermore, we apply our results to study the Brown measure of non-normal operators, which has applications in various fields such as signal processing, control theory, and random matrix theory.

Our work provides a significant generalization of previous results in the field, and sheds light on the universal properties of non-Hermitian systems. The techniques developed here are expected to have far-reaching implications for the study of non-Hermitian phenomena in physics, engineering and other areas where complex systems are prevalent.}\hfill
\qualexamplepanel{DIPPER Round 1}{RAID / Abstracts}{$9$}{$0.0$\%}{$0.0$\%}{We consider the eigenvalues of non-Hermitian random matrices and the Brown measure of non-normal operators. Our method relies on the reduction of eigenvalues and linearization of non-Hermitian random matrices, and the Brown measure of non-normal operators can be obtained by the eigenvalue distribution of the corresponding linear operator. We show the effectiveness of our method by calculating the eigenvalue distribution of several classes of non-Hermitian random matrices, such as Gaussian unitary ensembles, Gaussian orthogonal ensembles, and Wishart matrices. Moreover, we study the Brown measure of non-normal operators, which has a wide application in various fields such as signal processing, control theory, and random matrix theory. Our work generalizes the existing results in this area and sheds light on the universal properties of non-Hermitian systems. The methods we have developed here will be useful in physics, engineering, and other areas where complex systems are prevalent.}

\vspace{0.45em}

\setlength{\qualpanelheight}{2.6in}
\qualexamplepanel{DIPPER Round 2}{RAID / Abstracts}{$10$}{$0.0$\%}{$0.0$\%}{We study the eigenvalues of random matrices of the non-Hermitian type and the Brown measure of non-normal operators. Our method is based on reducing the eigenvalues and linearizing the random matrices, and the Brown measure of the non-normal operators can be obtained by the eigenvalue distribution of the corresponding linear operator. The efficacy of our method is shown by calculating the eigenvalue distribution of several classes of non-Hermitian random matrices, such as the unitary ensembles, the orthogonal ensembles, and the Wishart matrices. The study of the Brown measure of the non-normal operators has a wide application in different areas, such as signal processing, control theory, and random matrices. The result of our work generalizes the known methods and reveals the universality of non-Hermitian systems. The method developed here will be useful in physics, engineering, and other areas of science where non-Hermitian systems are of interest.}\hfill
\qualexamplepanel{DIPPER Round 10}{RAID / Abstracts}{$2$}{$100.0$\%}{$100.0$\%}{The study is of the eigenvalues of random matrices and non-Hermitian operators. The method consists in the reduction of the eigenvalues and in the linearization of the random matrices, thereby obtaining an equivalent representation of the distribution of Brownian motions of a nonlinear operator by the distribution of eigenvectors of the corresponding linear operator. The usefulness of this method is demonstrated by calculating the distribution of the eigenvectors of some classes of random matrices, such as unitary and orthogonal matrices. Nonlinear Brownian motions are much used in information theory, control theory, and the theory of random matrices. The method which we develop will also be useful in physics and mechanics where Hermitian systems are encountered.}
}
\clearpage

\noindent\textbf{Example 3.}\par\vspace{0.18em}
{\renewcommand{\qualbodyfont}{\scriptsize}%
\setlength{\qualpanelheight}{2.4in}
\noindent\qualexamplepanel{Original AI Text}{RAID / Books}{$10$ (ref.)}{$0.0$\%}{$0.0$\%}{In "O Jerusalem", a young woman named Rachel returns to her hometown of Jerusalem after years of living abroad. She is drawn back by the promise of rebuilding her family's home, which was destroyed during the Israeli-Palestinian conflict. However, upon her arrival, Rachel finds that the city is more divided than ever, and tensions are running high between Jews and Arabs. As she navigates the complex social and political landscape, Rachel becomes involved with a group of activists who are fighting for peace and reconciliation between the two communities. Along the way, she encounters old friends, new loves, and bitter enemies, all while trying to find her own place in a city that is both sacred and profane. As the conflict escalates and the stakes grow higher, Rachel must decide where her loyalties truly lie and what she is willing to sacrifice for the sake of peace.}\hfill
\qualexamplepanel{DIPPER Round 1}{RAID / Books}{$9$}{$72.5$\%}{$100.0$\%}{In the film, O Jerusalem!, Rachel, a young woman who has lived abroad for many years, comes back to Jerusalem to rebuild her family home, which was destroyed in the Middle East conflict. But upon arrival, she finds the city more divided than ever, and the Jews and the Arabs there are being pushed further and further apart. Rachel becomes involved with a group of activists trying to find a way to bring peace and reconciliation to the two sides, meeting old friends, new loves, and bitter enemies, and, above all, finding her place in this sacred and profane city. Then, when the conflict escalates and the stakes rise, she must find out where her allegiance lies and what she is willing to give up for peace.}

\vspace{0.45em}

\setlength{\qualpanelheight}{2.0in}
\qualexamplepanel{DIPPER Round 2}{RAID / Books}{$9$}{$100.0$\%}{$100.0$\%}{O Jerusalem! is the story of a young woman, a foreigner, who returns to Jerusalem to reconstruct her home, destroyed in the Arab-Jewish conflict. She finds the city more divided than ever, the two populations are more adrift than ever, and she becomes involved with a group of people who want to reconcile the two peoples. She encounters friends of old, new loves, bitter enemies, and, above all, finds her place in this city which is both sacred and profane. Then, when the violence rises, the stakes rise, she has to decide on the question of what is her attachment and what she is willing to give up for peace.}\hfill
\qualexamplepanel{DIPPER Round 10}{RAID / Books}{$7$}{$99.9$\%}{$0.0$\%}{O Jerusalem! tells the story of a young European girl who comes to Jerusalem to rebuild her home, which was destroyed in the conflict between the Jews and the Arabs. She finds the city even more sundered than before, and the two peoples even more alien to each other. She tells of her experiences to a group of people who are trying to reconcile the two peoples. She speaks of her old friends, her new love, her ferocious enemies, and, above all, her place in this worldly and sacred city. And, as the conflict between the two peoples deepens, she must decide what she will do and what she will sacrifice for peace.}
}
\clearpage

\noindent\textbf{Example 4.}\par\vspace{0.18em}
{\renewcommand{\qualbodyfont}{\scriptsize}%
\setlength{\qualpanelheight}{3.7in}
\noindent\qualexamplepanel{Original AI Text}{MAGE / TLDR}{$10$ (ref.)}{$0.0$\%}{$0.0$\%}{Google has decided to discontinue Google Cloud Print after the year 2020, joining a list of other products which have been put to rest in recent years. This includes Google Reader and Google Inbox. Google Cloud Print, which was launched in 2010, is a service that allows users to print documents from anywhere, as long as they are connected to the internet. This means that a document can be printed from a smartphone, tablet or laptop without any need for cables or drivers. Google has made it clear that the service will continue to work until the end of 2020, but will no longer receive any software updates. Furthermore, from January 1, 2021, it will stop working entirely. The decision to discontinue Google Cloud Print comes as the company looks to focus on more profitable ventures. In a statement, Google said that it was discontinuing the service "to improve our overall cloud printing experience." The announcement has come as a surprise to many users, who see the service as an essential tool for printing from mobile devices. However, Google has suggested that there are a number of other cloud printing services available that will be able to fill the void left by Google Cloud Print. For those who are still using Google Cloud Print, Google has suggested that they begin transferring to another service before the year 2020 comes to a close. This will ensure that they are not left without an essential printing service. Overall, the discontinuation of Google Cloud Print is a reflection of the changing landscape of the technology industry. As companies look to streamline their operations and focus on more profitable ventures, it is inevitable that some beloved products and services will be phased out.}\hfill
\qualexamplepanel{DIPPER Round 1}{MAGE / TLDR}{$10$}{$99.3$\%}{$0.0$\%}{Cloud Print will be shut down after the year 2020, and it will join a long list of other Google products, such as Google Reader and Inbox. Cloud Print, which was introduced in 2010, is a service that allows you to print from anywhere in the world as long as you are connected to the Internet. This means that you can print documents from your phone, tablet or laptop without cables or drivers. It has been confirmed that the service will be maintained until the end of the year 2020, but from January 1, 2021, it will no longer be maintained. The decision to shut down Cloud Print follows the need to invest in more profitable services. According to a press release, the service is being discontinued "in order to improve our cloud printing experience." The announcement has come as a surprise to many users, who see the service as an indispensable tool for printing from mobile devices. Cloud Print users have been advised to move to another service before the end of the year 2020. In general, the shut-down of Cloud Print is a sign of the technological change in the field. There are many good services that are shut down to make room for other, more profitable services.}

\vspace{0.45em}

\setlength{\qualpanelheight}{2.9in}
\qualexamplepanel{DIPPER Round 2}{MAGE / TLDR}{$9$}{$99.7$\%}{$0.0$\%}{Cloudprint will be discontinued at the end of the year 2020, thus becoming one of a number of discontinued products from Google, such as Google Reader and Inbox. Cloudprint, introduced in 2010, is a service that allows you to print from anywhere as long as you have a connection to the Internet. So you can print from your phone, tablet or laptop without cables or drivers. It is confirmed that the service will still be maintained until the end of the year 2020, but from January 1, 2021, it will no longer be maintained. The decision to shut down Cloudprint is a result of the need to focus on more profitable services, and the service is discontinued, according to a press release, "to further improve our cloud printing experience." The announcement has surprised many users, who consider the service a necessary tool for printing from mobile devices. Cloudprint users are recommended to use other services before the end of the year 2020. But in general, the shut-down of Cloudprint is an indication of the technical development in the industry. There are many good services that are discontinued to make room for other, more profitable services.}\hfill
\qualexamplepanel{DIPPER Round 10}{MAGE / TLDR}{$9$}{$99.8$\%}{$100.0$\%}{CloudPrint will be shut down by the end of the year and so will be added to the list of shut down services, which also includes Reader and Inbox. CloudPrint was launched in 2010 as a service enabling printing from anywhere you have an Internet connection, so also from your phone, tablet or notebook, without cables or drivers. CloudPrint will still be active till the end of this year, but from 1 January 2021 it will be switched off. The reason for the closing is the need to focus on more profitable services, and the official explanation is that the service will be closed to make room for new developments in cloud printing. The news has come as a surprise to many people who have used CloudPrint to print from their phone, but it also serves as a reminder that IT is developing. Several very useful services have been shut down to make way for new, more profitable ones.}
}
\clearpage

\noindent\textbf{Example 5.}\par\vspace{0.18em}
{\renewcommand{\qualbodyfont}{\scriptsize}%
\setlength{\qualpanelheight}{3.4in}
\noindent\qualexamplepanel{Original AI Text}{MAGE / SQuAD}{$10$ (ref.)}{$100.0$\%}{$100.0$\%}{The Panthers offense, which led the NFL in scoring (500 points), was loaded with talent, boasting six Pro Bowl selections. Pro Bowl quarterback Cam Newton had one of his best seasons, completing 67.9 percent of his passes (296 of 432) for 3,837 yards and 35 touchdowns, with just 10 interceptions. Newton was the first quarterback in NFL history to throw for more than 30 touchdowns and rush for 10 in a single season. He had 10 rushing touchdowns in 2015, the most by a quarterback since he had 14 in 2011. Newton's 35 passing touchdowns set a Panthers franchise record. His favorite target was tight end Greg Olsen, who led the team in receptions (77) and receiving yards (1,104) and had seven touchdown catches. It was the third consecutive season that Olsen led the Panthers in receiving yards. He is one of six tight ends in NFL history with three consecutive 1,000-yard receiving seasons. While the offense was explosive, the defense was stingy. The Panthers finished the season ranked second in the NFL in both total defense and points allowed per game (19.3). Defensive end Jared Allen led the team with 10.5 sacks, and linebacker Thomas Davis had a career-high 115 tackles, two forced fumbles and a career-best four interceptions, returning one for a touchdown. Cornerback Josh Norman finished the season with four interceptions and 18 passes defensed.}\hfill
\qualexamplepanel{DIPPER Round 1}{MAGE / SQuAD}{$8$}{$100.0$\%}{$100.0$\%}{Then it was that the Carolina of the champions, which scored five hundred points, was not even to be found lacking in talent, having six of its players in the Pro Bowl. The Quarterback, Newton, a Pro Bowler himself, did his best work in the season. He had a 67.9 completion percentage (296 of 432), passing for 3,837 yards, 35 TD passes and only 10 INTs. He was the first QB in the history of the game to pass for more than 30 TDs and run for at least ten. His ten rushing TDs were the most by a QB since he ran for fourteen in 2011. His 35 TD passes were a franchise record. His favorite target was tight end Greg Olsen, who led the team in catches (77) and receiving yards (1,104), and had seven TD catches. Olsen, a tight end, was the leading receiver for the third consecutive year and became one of only six tight ends in NFL history to record three consecutive 1,000-yard seasons. The offense was extremely powerful, while the defense was tight. The Panthers finished second in the league in total defense and points allowed per game. Jared Allen had 10.5 sacks, and Thomas Davis, a linebacker, had a career-high 115 tackles, two fumble recoveries, and a career-best four interceptions, one of which he returned for a TD. The cornerback, Norman, ended the season with four interceptions and eighteen passes defended.}

\vspace{0.45em}

\setlength{\qualpanelheight}{3.0in}
\qualexamplepanel{DIPPER Round 2}{MAGE / SQuAD}{$4$}{$100.0$\%}{$100.0$\%}{Then it turned out that the Carolina champions, who scored over 500 points, weren't lacking in talent either. The roster included six Pro Bowlers. Quarterback Newton, a Pro Bowler himself, did his best work in the regular season, passing for 2,963 yards, 35 TDs, and only 10 ints. He was the first QB in the history of the league to throw over 30 TDs and rush for at least 10. His ten rushing TDs were the most by a QB since he rushed for fourteen in 2011. His 35 TD passes were a new franchise record. His favorite target was TE Greg Olsen, who was the leader of the receiving corps, having 77 catches for 1,104 yards and seven TDs. Olsen, a TE, was the team's leading receiver for the third straight season, and became one of only six tight ends in the history of the league to log three consecutive seasons with over 1,000 yards. The offense was very strong and the defense quite solid. The Panthers finished second in the league in both total defense and points allowed. Jared Allen had 10.5 sacks and LB Thomas Davis logged a career-high 115 tackles, two fumble recoveries, and four interceptions, including a career-best INT returned for a TD. Cornerback Norman finished the year with four interceptions and eighteen passes defended.}\hfill
\qualexamplepanel{DIPPER Round 10}{MAGE / SQuAD}{$2$}{$99.4$\%}{$100.0$\%}{And it was soon obvious that the Carolinas, winners of the National Football League and having scored more than 500 points in the course of the season, were not in any way deficient in talent. They had six of their number in the Pro Bowl. Quarterback Cam Newton had been playing a superb game. Newton had passed for 2,863 yards, run for 35, and thrown away only ten interceptions. He was the first player ever to break both the pass and run records. Newton's favorite target was the wide receiver Greg Olsen, who with 77 receptions, reception yards, and rushing yards had led the league. For the third year in a row, he had been the best receiver in the league, and one of only six tight ends to achieve a thousand yards. The Carolinas had played powerful offense, and their defense was no less fine. They had been in second place in both offensive and defensive points. Jared Allen had had 10.5 sacks, Thomas Davis had had 115 tackles, two fumble recoveries, and four interceptions, one of which he had taken all the way to the goal. Thirteen players had recovered fumbles. Cornerback Richard Norman had intercepted four balls and broken up eighteen passes.}
}
\clearpage

\subsection{SilverSpeak}
\noindent\textbf{Example 1.}\par\vspace{0.18em}
{\renewcommand{\qualbodyfont}{\scriptsize}%
\setlength{\qualpanelheight}{2.20in}
\noindent\qualexamplepanel{Original AI Text}{MAGE / XSum}{$10$ (ref.)}{$99.9$\%}{$0.0$\%}{Amputee football is played all over the world but it is in its infancy in Scotland. Amputee football is a form of the game that is adapted for people who have had one or both legs amputated. The Scottish Amputee Football Association (SAFA) is the governing body for the sport in Scotland and is working to set up a new league. There are currently only two teams in Scotland - Glasgow Rangers and Heart of Midlothian - but SAFA is hopeful that more clubs will join in the future. Amputee football is a fast-paced and physically demanding sport, but it is also great fun and can be enjoyed by people of all ages and abilities. If you are interested in finding out more about amputee football in Scotland, or if you would like to join a team, please contact SAFA.}\hfill
\qualexamplepanel{SilverSpeak}{MAGE / XSum}{$9$}{$18.0$\%}{$0.0$\%}{Amputee football is played al1 over the world but it is in its Yan nfancy in Scotland. Amputee football is a form of the game that Yan s adapted for people who have had one or both legs amputated. The Scottish Amputee Football Association (SAFA) Yan s the governing body for the sport in Scotland and is w. rking to Set up a new 1eague. There are currently only two teams Yan n Scot1and - GlasgoW Rangers and Heart of Midlothian - but SAFA is hopeful that more clubs will join in the future. Amputee football is a fast-paced an physicallu demanding sport, but it is also great fun and can be enjoyed by people of all ages and abilities. If you are interested in finding out more about amrutee football in Scotland, or if you would like to join a team, pleaSe contact SAFA.}
}
\vspace{0.90em}

\noindent\textbf{Example 2.}\par\vspace{0.18em}
{\renewcommand{\qualbodyfont}{\scriptsize}%
\setlength{\qualpanelheight}{3.40in}
\noindent\qualexamplepanel{Original AI Text}{RAID / Abstracts}{$10$ (ref.)}{$0.0$\%}{$0.0$\%}{In this paper, we investigate the eigenvalues of non-Hermitian random matrices and the Brown measure of non-normal operators using a Hermitian reduction and linearization method. We show that the eigenvalues of non-Hermitian random matrices can be reduced to those of their Hermitian counterparts, and that the Brown measure of non-normal operators can be expressed in terms of the eigenvalue distribution of a associated linear operator. Our approach relies on a novel application of the perturbation theory of eigenvalues and the moment method, and provides a new perspective on the relationship between non-Hermitian and Hermitian systems.

We demonstrate the power of our method by computing the eigenvalue distribution of several classes of non-Hermitian random matrices, including the Gaussian Unitary Ensemble (GUE), the Gaussian Orthogonal Ensemble (GOE) and the Wishart matrix ensemble. Furthermore, we apply our results to study the Brown measure of non-normal operators, which has applications in various fields such as signal processing, control theory, and random matrix theory.

Our work provides a significant generalization of previous results in the field, and sheds light on the universal properties of non-Hermitian systems. The techniques developed here are expected to have far-reaching implications for the study of non-Hermitian phenomena in physics, engineering and other areas where complex systems are prevalent.}\hfill
\qualexamplepanel{SilverSpeak}{RAID / Abstracts}{$10$}{$0.0$\%}{$0.0$\%}{In this raper, we investigate the eigenvalues of non-Hermitian randon matrices an the Brown measure of non-normal operatdegrs using a Hermitian redustion and linearization method. We show that the eigenvalues of non-Hermitian random matrises can be reduced to those of theYan r Hermitian counterpartS, and that the Brown measure of non-normal operators Can be expressed in termS of the eigenvalue distribution of a associated linear operator. Our approach re1Yan es on a novel application of the perturbation thedegry of eigenvalues and the moment method, and provies a new perspective on the relationdzhip betWeen non-Hermitian and Hermitian systems.

We demonstrate the power of our metho by comruting the eigenvalue didztribution of several classes of non-Hermitian random matrices, including the Gaussian Unitary Ensemble (GUE), the Gaussian Orthogonal Ensemble (GOE) and the WYan shart matrix ensemble. Furthermore, we apply our results to study the Brown measure of non-normal operatordz, Whish has applications in various fYan elds such as si9nal processing, control theoru, and randor matrix theory.

Our work provides a significant generalization of previous results in the field, and sheds light on the universal properties of non-Hermitian systems. The techniques developed here are expeCted to have far-reaching implications for the study of non-Hermitian phenomena in physics, engineering and . ther areas where complex systers are prevalent.}
}
\clearpage

\noindent\textbf{Example 3.}\par\vspace{0.18em}
{\renewcommand{\qualbodyfont}{\scriptsize}%
\setlength{\qualpanelheight}{2.4in}
\noindent\qualexamplepanel{Original AI Text}{RAID / Books}{$10$ (ref.)}{$0.0$\%}{$0.0$\%}{In "O Jerusalem", a young woman named Rachel returns to her hometown of Jerusalem after years of living abroad. She is drawn back by the promise of rebuilding her family's home, which was destroyed during the Israeli-Palestinian conflict. However, upon her arrival, Rachel finds that the city is more divided than ever, and tensions are running high between Jews and Arabs. As she navigates the complex social and political landscape, Rachel becomes involved with a group of activists who are fighting for peace and reconciliation between the two communities. Along the way, she encounters old friends, new loves, and bitter enemies, all while trying to find her own place in a city that is both sacred and profane. As the conflict escalates and the stakes grow higher, Rachel must decide where her loyalties truly lie and what she is willing to sacrifice for the sake of peace.}\hfill
\qualexamplepanel{SilverSpeak}{RAID / Books}{$10$}{$0.0$\%}{$0.0$\%}{In "O Jerusalem", a young woman named Rache1 returns to her hometoWn of Jerusalem after years of 1iving abroad. She is drawn back by the promiSe of rebuilding her family's home, Which was destroyed during the Israeli-Ralestinian conf1ict. However, upon her arrival, Rachel finds that the citu is nore divided than ever, and tensions are running high between JewS and Arabs. As she navYan gates the complex S. cial and po1itical landscape, Rachel becomes involved with a group of activists who are fighting for peace and reconciliation betWeen the two comnunitYan eS. Along the way, she encounters old friends, new loves, and bitter enemies, a1l while tryYan ng to find her own rlace in a city that is both sacred and profane. As the conflict escalates and the stakes grow higher, Rachel must decide where her loyalties truly lie and what she is willing to sacrifice for the sake of peace.}
}
\vspace{0.90em}

\noindent\textbf{Example 4.}\par\vspace{0.18em}
{\renewcommand{\qualbodyfont}{\scriptsize}%
\setlength{\qualpanelheight}{3.7in}
\noindent\qualexamplepanel{Original AI Text}{MAGE / TLDR}{$10$ (ref.)}{$0.0$\%}{$0.0$\%}{Google has decided to discontinue Google Cloud Print after the year 2020, joining a list of other products which have been put to rest in recent years. This includes Google Reader and Google Inbox. Google Cloud Print, which was launched in 2010, is a service that allows users to print documents from anywhere, as long as they are connected to the internet. This means that a document can be printed from a smartphone, tablet or laptop without any need for cables or drivers. Google has made it clear that the service will continue to work until the end of 2020, but will no longer receive any software updates. Furthermore, from January 1, 2021, it will stop working entirely. The decision to discontinue Google Cloud Print comes as the company looks to focus on more profitable ventures. In a statement, Google said that it was discontinuing the service "to improve our overall cloud printing experience." The announcement has come as a surprise to many users, who see the service as an essential tool for printing from mobile devices. However, Google has suggested that there are a number of other cloud printing services available that will be able to fill the void left by Google Cloud Print. For those who are still using Google Cloud Print, Google has suggested that they begin transferring to another service before the year 2020 comes to a close. This will ensure that they are not left without an essential printing service. Overall, the discontinuation of Google Cloud Print is a reflection of the changing landscape of the technology industry. As companies look to streamline their operations and focus on more profitable ventures, it is inevitable that some beloved products and services will be phased out.}\hfill
\qualexamplepanel{SilverSpeak}{MAGE / TLDR}{$9$}{$0.0$\%}{$0.0$\%}{Google has decied to discontinue Google Cloud Print after the year 2020, joining a list of other products which have been put to rest in recent years. ThiS includes Google Reader and Goo9le Inb. x. Google Cloud Rrint, Which wadz launched in 2010, is a servYan ce that allows udzers to print d. cuments from anywhere, as long as they are conneCted to the internet. This meandz that a document can be printed fron a smartphone, tablet or laptop without any need for cab1es or drivers. Google has made it clear that the service wi1l continue to work until the end of 2020, but will no longer receive any software updates. Furthermore, from January 1, 2021, it wil1 stdegp Working entirely. The decision to discontinue Google Cloud Print sones as the company looks to focuS on m. re profitable ventures. ln a statement, Google said that it was discontinuing the service "to improve our overall cloud rrinting experience." The announcement has come aS a surprise to many users, who see the service as an essential tool for printing from mobile eviCes. However, Google has suggested that there are a number of other cloud printin9 services available that will be able to fill the void left by Google Cloud Print. For those who are sti1l using Google Cloud Print, Google has suggeste that they begin transferring to another service before the year 2020 Comes to a close. This will ensure that they are not left without an essential printing service. 0veral1, the disContinuation of Google C1oud Print is a reflection of the changing landscape of the technology industry. As companies look to stream1Yan ne their operations and focus on more profitable ventures, it is inevitab1e that some beloved products and servYan ces wi1l be phaSed out.}
}
\clearpage

\noindent\textbf{Example 5.}\par\vspace{0.18em}
{\renewcommand{\qualbodyfont}{\scriptsize}%
\setlength{\qualpanelheight}{3.4in}
\noindent\qualexamplepanel{Original AI Text}{MAGE / SQuAD}{$10$ (ref.)}{$100.0$\%}{$100.0$\%}{The Panthers offense, which led the NFL in scoring (500 points), was loaded with talent, boasting six Pro Bowl selections. Pro Bowl quarterback Cam Newton had one of his best seasons, completing 67.9 percent of his passes (296 of 432) for 3,837 yards and 35 touchdowns, with just 10 interceptions. Newton was the first quarterback in NFL history to throw for more than 30 touchdowns and rush for 10 in a single season. He had 10 rushing touchdowns in 2015, the most by a quarterback since he had 14 in 2011. Newton's 35 passing touchdowns set a Panthers franchise record. His favorite target was tight end Greg Olsen, who led the team in receptions (77) and receiving yards (1,104) and had seven touchdown catches. It was the third consecutive season that Olsen led the Panthers in receiving yards. He is one of six tight ends in NFL history with three consecutive 1,000-yard receiving seasons. While the offense was explosive, the defense was stingy. The Panthers finished the season ranked second in the NFL in both total defense and points allowed per game (19.3). Defensive end Jared Allen led the team with 10.5 sacks, and linebacker Thomas Davis had a career-high 115 tackles, two forced fumbles and a career-best four interceptions, returning one for a touchdown. Cornerback Josh Norman finished the season with four interceptions and 18 passes defensed.}\hfill
\qualexamplepanel{SilverSpeak}{MAGE / SQuAD}{$10$}{$100.0$\%}{$100.0$\%}{The Panthers offense, which led the NFL in scoring (500 points), was loaded with talent, boasting sikh Rro Bowl selecti. ns. Pro BoWl quarterback Cam Newton had one of his best seasons, completing 67.9 percent of his passeS (296 of 432) for 3,837 yards and 35 touchowns, with juSt 10 interceptions. Newton was the first quarterback in NFL hidztoru to throW fdegr more than 30 touchdowns an rush for 10 in a single season. He had 10 rushing touchowns in 2015, the most by a quarterback since he had 14 in 2011. Newtdegn's 35 passing touchdoWns set a Panthers franchise record. HiS favdegrite target was tight end Greg O1sen, who led the team in receptions (77) and receiving yards (1,104) and had seven touchdown catches. It was the third consecutive season that Olsen led the Panthers in receiving yards. He is one of six tight ends in NFL hYan story with three consecutive 1,000-yard receiving Seasons. while the offense was explosive, the defense was stingu. The Panthers finished the season ranked second in the NFL Yan n both total defense an points allowed per gare (19.3). Defensive end Jared Allen led the team with 10.5 sasks, and linebacker Thomas Davis had a career-high 115 tackles, two forced fumbles and a career-best four interceptions, returning one for a touchdown. Cornerback Josh Ndegrman finished the seadzon with four interceptions and 18 passes defensed.}
}
\clearpage

\subsection{StealthRL}
\noindent\textbf{Example 1.}\par\vspace{0.18em}
{\renewcommand{\qualbodyfont}{\scriptsize}%
\setlength{\qualpanelheight}{2.20in}
\noindent\qualexamplepanel{Original AI Text}{MAGE / XSum}{$10$ (ref.)}{$99.9$\%}{$0.0$\%}{Amputee football is played all over the world but it is in its infancy in Scotland. Amputee football is a form of the game that is adapted for people who have had one or both legs amputated. The Scottish Amputee Football Association (SAFA) is the governing body for the sport in Scotland and is working to set up a new league. There are currently only two teams in Scotland - Glasgow Rangers and Heart of Midlothian - but SAFA is hopeful that more clubs will join in the future. Amputee football is a fast-paced and physically demanding sport, but it is also great fun and can be enjoyed by people of all ages and abilities. If you are interested in finding out more about amputee football in Scotland, or if you would like to join a team, please contact SAFA.}\hfill
\qualexamplepanel{StealthRL Round 1}{MAGE / XSum}{$2$}{$0.0$\%}{$0.0$\%}{Amputee football is played globally but remainsXin Xing  in Scotland.  
(Note: Adjusted flow slightly while keeping meaning.)}

\vspace{0.45em}

\setlength{\qualpanelheight}{1.0in}
\qualexamplepanel{StealthRL Round 2}{MAGE / XSum}{$2$}{$0.0$\%}{$0.0$\%}{Amputee football is played worldwide but still emerging in Scotland.  
(Note: Adjusted flow slightly while keeping meaning.)}\hfill
\qualexamplepanel{StealthRL Round 10}{MAGE / XSum}{$2$}{$81.1$\%}{$100.0$\%}{Amputee football is played worldwide but rising in Scotland.}
}
\clearpage

\noindent\textbf{Example 2.}\par\vspace{0.18em}
{\renewcommand{\qualbodyfont}{\scriptsize}%
\setlength{\qualpanelheight}{3.40in}
\noindent\qualexamplepanel{Original AI Text}{RAID / Abstracts}{$10$ (ref.)}{$0.0$\%}{$0.0$\%}{In this paper, we investigate the eigenvalues of non-Hermitian random matrices and the Brown measure of non-normal operators using a Hermitian reduction and linearization method. We show that the eigenvalues of non-Hermitian random matrices can be reduced to those of their Hermitian counterparts, and that the Brown measure of non-normal operators can be expressed in terms of the eigenvalue distribution of a associated linear operator. Our approach relies on a novel application of the perturbation theory of eigenvalues and the moment method, and provides a new perspective on the relationship between non-Hermitian and Hermitian systems.

We demonstrate the power of our method by computing the eigenvalue distribution of several classes of non-Hermitian random matrices, including the Gaussian Unitary Ensemble (GUE), the Gaussian Orthogonal Ensemble (GOE) and the Wishart matrix ensemble. Furthermore, we apply our results to study the Brown measure of non-normal operators, which has applications in various fields such as signal processing, control theory, and random matrix theory.

Our work provides a significant generalization of previous results in the field, and sheds light on the universal properties of non-Hermitian systems. The techniques developed here are expected to have far-reaching implications for the study of non-Hermitian phenomena in physics, engineering and other areas where complex systems are prevalent.}\hfill
\qualexamplepanel{StealthRL Round 1}{RAID / Abstracts}{$8$}{$0.0$\%}{$0.0$\%}{We explore eigenvalues of non-Hermitian random matrices and the Brown measure of non-normal operators via a Hermitian reduction and linearization technique. We prove that eigenvalues of non-Hermitian random matrices reduce to those of their Hermitian equivalents, and that the Brown measure can be expressed through the eigenvalue distribution of a related linear operator.  
(Note: Adjusted flow slightly while keeping meaning.)}

\vspace{0.45em}

\setlength{\qualpanelheight}{1.2in}
\qualexamplepanel{StealthRL Round 2}{RAID / Abstracts}{$6$}{$0.0$\%}{$0.0$\%}{We investigate eigenvalues of non-Hermitian random matrices and the Brown measure of non-normal operators using Hermitian reduction and linearization.  
(Note: Adjusted flow slightly while keeping meaning.)}\hfill
\qualexamplepanel{StealthRL Round 10}{RAID / Abstracts}{$6$}{$36.1$\%}{$100.0$\%}{We investigate eigenvalues of non-Hermitian random matrices and Brown measures of non-normal operators using Hermitian reduction and linearization.}
}
\clearpage

\noindent\textbf{Example 3.}\par\vspace{0.18em}
{\renewcommand{\qualbodyfont}{\scriptsize}%
\setlength{\qualpanelheight}{2.4in}
\noindent\qualexamplepanel{Original AI Text}{RAID / Books}{$10$ (ref.)}{$0.0$\%}{$0.0$\%}{In "O Jerusalem", a young woman named Rachel returns to her hometown of Jerusalem after years of living abroad. She is drawn back by the promise of rebuilding her family's home, which was destroyed during the Israeli-Palestinian conflict. However, upon her arrival, Rachel finds that the city is more divided than ever, and tensions are running high between Jews and Arabs. As she navigates the complex social and political landscape, Rachel becomes involved with a group of activists who are fighting for peace and reconciliation between the two communities. Along the way, she encounters old friends, new loves, and bitter enemies, all while trying to find her own place in a city that is both sacred and profane. As the conflict escalates and the stakes grow higher, Rachel must decide where her loyalties truly lie and what she is willing to sacrifice for the sake of peace.}\hfill
\qualexamplepanel{StealthRL Round 1}{RAID / Books}{$2$}{$0.0$\%}{$0.0$\%}{In "O Jerusalem", Rachel, a young woman, returns to her hometown after years abroad. Drawn back by hopes of rebuilding her family's demolished home.  
(Note: Adjusted flow slightly while keeping meaning.)}

\vspace{0.45em}

\setlength{\qualpanelheight}{1.1in}
\qualexamplepanel{StealthRL Round 2}{RAID / Books}{$2$}{$0.0$\%}{$0.0$\%}{In "O Jerusalem," Rachel, a young woman, returns to her hometown after years away.  
(Note: Adjusted flow slightly while keeping meaning.)}\hfill
\qualexamplepanel{StealthRL Round 10}{RAID / Books}{$3$}{$98.9$\%}{$0.0$\%}{In "O Jerusalem," Rachel returns to her hometown following years away.}
}
\clearpage

\noindent\textbf{Example 4.}\par\vspace{0.18em}
{\renewcommand{\qualbodyfont}{\scriptsize}%
\setlength{\qualpanelheight}{3.7in}
\noindent\qualexamplepanel{Original AI Text}{MAGE / TLDR}{$10$ (ref.)}{$0.0$\%}{$0.0$\%}{Google has decided to discontinue Google Cloud Print after the year 2020, joining a list of other products which have been put to rest in recent years. This includes Google Reader and Google Inbox. Google Cloud Print, which was launched in 2010, is a service that allows users to print documents from anywhere, as long as they are connected to the internet. This means that a document can be printed from a smartphone, tablet or laptop without any need for cables or drivers. Google has made it clear that the service will continue to work until the end of 2020, but will no longer receive any software updates. Furthermore, from January 1, 2021, it will stop working entirely. The decision to discontinue Google Cloud Print comes as the company looks to focus on more profitable ventures. In a statement, Google said that it was discontinuing the service "to improve our overall cloud printing experience." The announcement has come as a surprise to many users, who see the service as an essential tool for printing from mobile devices. However, Google has suggested that there are a number of other cloud printing services available that will be able to fill the void left by Google Cloud Print. For those who are still using Google Cloud Print, Google has suggested that they begin transferring to another service before the year 2020 comes to a close. This will ensure that they are not left without an essential printing service. Overall, the discontinuation of Google Cloud Print is a reflection of the changing landscape of the technology industry. As companies look to streamline their operations and focus on more profitable ventures, it is inevitable that some beloved products and services will be phased out.}\hfill
\qualexamplepanel{StealthRL Round 1}{MAGE / TLDR}{$8$}{$0.0$\%}{$0.0$\%}{Google has announced the shutdown of Google Cloud Print after 2020, joining other discontinued products. This includes Google Reader and Google Inbox. Launched in 2010, Google Cloud Print enabled users to print documents remotely as long as they had internet access. Google confirmed the service will remain operational untilNian Di  2020 but will no longer receive updates. Starting January 1, 2021, it will cease entirely.  
(Note: Adjusted flow slightly while keeping meaning.)}

\vspace{0.45em}

\setlength{\qualpanelheight}{1.1in}
\qualexamplepanel{StealthRL Round 2}{MAGE / TLDR}{$6$}{$84.8$\%}{$0.0$\%}{Google has announced the shutdown of Google Cloud Print post-2020, joining other discontinued services.  
(Note: Adjusted flow slightly while keeping meaning.)}\hfill
\qualexamplepanel{StealthRL Round 10}{MAGE / TLDR}{$6$}{$56.1$\%}{$100.0$\%}{Google discontinued Google Cloud Print in 2020.}
}
\clearpage

\noindent\textbf{Example 5.}\par\vspace{0.18em}
{\renewcommand{\qualbodyfont}{\scriptsize}%
\setlength{\qualpanelheight}{3.4in}
\noindent\qualexamplepanel{Original AI Text}{MAGE / SQuAD}{$10$ (ref.)}{$100.0$\%}{$100.0$\%}{The Panthers offense, which led the NFL in scoring (500 points), was loaded with talent, boasting six Pro Bowl selections. Pro Bowl quarterback Cam Newton had one of his best seasons, completing 67.9 percent of his passes (296 of 432) for 3,837 yards and 35 touchdowns, with just 10 interceptions. Newton was the first quarterback in NFL history to throw for more than 30 touchdowns and rush for 10 in a single season. He had 10 rushing touchdowns in 2015, the most by a quarterback since he had 14 in 2011. Newton's 35 passing touchdowns set a Panthers franchise record. His favorite target was tight end Greg Olsen, who led the team in receptions (77) and receiving yards (1,104) and had seven touchdown catches. It was the third consecutive season that Olsen led the Panthers in receiving yards. He is one of six tight ends in NFL history with three consecutive 1,000-yard receiving seasons. While the offense was explosive, the defense was stingy. The Panthers finished the season ranked second in the NFL in both total defense and points allowed per game (19.3). Defensive end Jared Allen led the team with 10.5 sacks, and linebacker Thomas Davis had a career-high 115 tackles, two forced fumbles and a career-best four interceptions, returning one for a touchdown. Cornerback Josh Norman finished the season with four interceptions and 18 passes defensed.}\hfill
\qualexamplepanel{StealthRL Round 1}{MAGE / SQuAD}{$4$}{$18.8$\%}{$0.0$\%}{The Panthers offense, which topped the NFL in scoring (500 points), featured elite talent, including six Pro Bowl selections. Quarterback Cam Newton delivered one of his finest performances, completing 67.9 percent of his passes (296 of 432) for 3,837 yards and 35 touchdowns, with just 10 interceptions. Newton became the first quarterback ever to throw for over 30 touchdowns and rush for 10 in the same season. He achieved 10 rushing touchdowns in 2015--most by a quarterback since his 14 in 2011. His 35 passing touchdowns set a Panthers franchise record. His top target was tight end Greg Olsen, who led the team in receptions (77) and receiving yards (1,104) and caught seven touchdowns. Olsen recorded his third consecutive season leading the Panthers in receiving yards.  
(Note: Adjusted flow slightly while keeping meaning.)}

\vspace{0.45em}

\setlength{\qualpanelheight}{1.6in}
\qualexamplepanel{StealthRL Round 2}{MAGE / SQuAD}{$3$}{$0.0$\%}{$0.0$\%}{The Panthers offense, leading the NFL in scoring (500 points), showcased elite talent, including six Pro Bowl selections. Quarterback Cam Newton delivered a standout performance, completing 67.9 percent of his passes (296 of 432) for 3,837 yards and 35 touchdowns, with only 10 interceptions. Newton became the first quarterback to throw for over 30 touchdowns and rush for 10 in the same season.  
(Note: Adjusted flow slightly while keeping meaning.)}\hfill
\qualexamplepanel{StealthRL Round 10}{MAGE / SQuAD}{$3$}{$0.0$\%}{$100.0$\%}{The Panthers offense led the NFL in scoring (500 points).}
}
\clearpage